\documentclass{article}

\usepackage{geometry}
\usepackage{graphicx}%
\usepackage{multirow}%
\usepackage{amsmath,amssymb,amsfonts}%
\usepackage{amsthm}%
\usepackage{mathrsfs}%
\usepackage[title]{appendix}%
\usepackage{xcolor}%
\usepackage{textcomp}%
\usepackage{manyfoot}%
\usepackage{booktabs}%
\usepackage{algorithm}%
\usepackage{algorithmicx}%
\usepackage{algpseudocode}%
\usepackage{listings}%
\usepackage{url}
\usepackage{authblk}

\providecommand{\keywords}[1]
{
  \small	
  \textbf{\textit{Keywords---}} #1
}

\renewcommand\footnotetext[2][]{{\removelastskip\vskip2mm%
\let
\hskip0\if!##1!\else{\smash{$^{#1}$}}#2\par}}%

\begin{document}

\title{Cost-Sensitive Evaluation for Binary Classifiers}

\author[1]{Pierangelo Lombardo}
\author[1]{Antonio Casoli}
\author[2]{Cristian Cingolani}
\author[2]{Shola Oshodi}
\author[2]{Michele Zanatta}

\affil[1]{Eutelsat, Paris, France}
\affil[2]{Reply, Rome, Italy}

\date{}

\maketitle

\begin{abstract}
Selecting an appropriate evaluation metric for classifiers is crucial for model comparison, parameter optimization, and deployment decisions, yet there is no consensus on a broadly accepted evaluation paradigm explicitly aligned with Total Classification Cost (TCC) minimization. At the same time, class imbalance is often treated as a problem to be corrected \emph{per se}, potentially causing misalignments with TCC minimization.

To address these limitations, (\emph{i}) we define Weighted Accuracy (WA), an evaluation metric for binary classifiers with a straightforward interpretation as a weighted version of accuracy and 
(\emph{ii}) we propose a general reweighting framework for handling class imbalance in cost-sensitive scenarios, providing an alternative to resampling techniques.
This framework applies to any evaluation metric or loss function
that can be expressed as a linear combination of example-dependent quantities; it enables meaningful comparison of evaluation results obtained on different datasets and accounts for discrepancies between the \emph{development} dataset, used for training, validation, and testing, and the \emph{target} dataset, where the model will be deployed.
Within this framework, we derive the conditions under which standard rebalancing techniques remain coherent with TCC minimization, and when they may instead become misleading.

We prove that, under example-independent Unit Classification Costs, maximizing WA is equivalent to minimizing TCC.
Finally, we analyze the robustness of WA in realistic example-dependent cost scenarios by studying its correlation with TCC across a broad range of class imbalance and cost regimes. The results show that WA maintains robust alignment with TCC across almost all examined scenarios.
\end{abstract}


\section{Introduction}\label{sec1}

Selecting the evaluation metric for a Machine Learning (ML) use case is a fundamental step, as it guides the choice of the optimal model and its hyperparameters \cite{new_metrics2009measuring,metrics2011evaluation,new_metrics2012novel,metrics2015review,metrics2023metrics, metrics2024evaluation}. 
However, this task is still complex: numerous metrics exist, and they can rank models in completely different ways \cite{metrics2014strategy}, with no consensus on the optimal choice \cite{metrics2015review, metrics2011evaluation, new_metrics2012novel,metrics2023metrics, metrics2024evaluation, new_metrics2024f_1}.
In classification tasks, model performance is typically described using the confusion matrix and evaluated using derived metrics such as accuracy, F-measure, Receiver Operating Characteristic-Area Under the Curve (ROC-AUC), and Cohen’s kappa~\cite{metrics2015review, metrics2011evaluation, new_metrics2022extending, metrics2024evaluation, new_metrics2024f_1}. Changing the evaluation metric often leads to selecting a different optimal classifier.
This is related to the fact that these metrics implicitly define the Unit Classification Costs (UCCs), i.e., the cost of classifying an example of class $i$ as class $j$, in an uncontrolled and potentially incorrect manner. For example, in binary classification, 
accuracy implies equal costs for false positives and false negatives, while ROC-AUC corresponds to variable cost distributions across classifiers \cite{new_metrics2009measuring}.

Moreover, most real-world datasets are moderately or heavily imbalanced: this is common in domains such as medical diagnosis \cite{cost_sensitive2006influence,cost_sensitive2014imbalanced,imbalance2017imbalanced,imbalance2020streams}, biochemical forecasting (e.g., protein sequences, gene expression) \cite{imbalance2017imbalanced,new_metrics2019learning}, image processing \cite{metrics2024evaluation}, intrusion detection \cite{new_metrics2006b,cost_sensitive2006influence}, fraud detection (e.g., banking, telecommunications) \cite{cost_sensitive2006influence,cost_sensitive2014imbalanced,imbalance2017imbalanced,imbalance2020streams}, telecommunications applications \cite{imbalance2010data,imbalance2017imbalanced}, information retrieval \cite{cost_sensitive2014imbalanced}, and anomaly detection in manufacturing \cite{cost_sensitive2014imbalanced}. 
In such cases, many standard metrics can distort results, making model ranking even more challenging \cite{imbalance2010data,new_metrics2019learning,new_metrics2022extending}. 
Class imbalance is widely recognized as one of the major challenges in ML research~\cite{cost_sensitive2014imbalanced, imbalance2017imbalanced,imbalance2020streams,imbalance2021performance}
 and several approaches have been proposed to mitigate its effects,
including undersampling, oversampling, hybrid sampling strategies (e.g., retaining difficult-to-learn samples), and Cost-Sensitive Machine Learning (CSML)~\cite{imbalance2002smote, imbalance2003c4, new_metrics2006b, imbalance2010data, cost_sensitive2014imbalanced, new_metrics2019metric, new_metrics2020imbalance, imbalance2020streams, imbalance2021balancing}.

CSML involves explicitly defining a cost matrix 
whose entries are the UCCs, in contrast with traditional machine learning, where such costs are only implicitly defined through the evaluation metric. 
Empirical studies have shown that when UCCs are explicitly estimated, they often exhibit substantial imbalances across real-world use cases such as medical diagnosis~\cite{cost_sensitive2006test, diabetes2007using, diabetes2020prediction}, bankruptcy prediction~\cite{cost_sensitive1977zetatm, cost_sensitive2000comparison}, loan credit rating~\cite{cost_sensitive2021multi}, cybersecurity~\cite{new_metrics2024f_1}, biometric identity recognition \cite{new_metrics2006mean}, consumer credit scoring~\cite{new_metrics2006mean}, and database marketing~\cite{cost_sensitive1999metacost}.
In business applications, a concept closely tied to the reduction of total cost in machine learning is Return on Investment (RoI). A similar principle applies in healthcare, where costs may include not only economic expenses but also health risks or limited availability of treatments \cite{cost_sensitive2006test, diabetes2007using, diabetes2020prediction}.

Within the umbrella of CSML, two distinct research directions exist:  
(i) studies that aim to estimate and minimize the Total Classification Cost (TCC)~\cite{cost_sensitive1977zetatm, cost_sensitive2006test, cost_matrix2012classification, imbalance2021performance}, which is the key for RoI optimization, and  
(ii) studies that use misclassification costs primarily to address class imbalance, often with the goal of maximizing specific metrics (e.g., accuracy, F1, specificity, sensitivity) on a rebalanced dataset~\cite{cost_sensitive1998cost, cost_sensitive2006influence, cost_sensitive2008cost, cost_sensitive2014imbalanced, imbalance2018curse, diabetes2020prediction, imbalance2020streams, cost_sensitive2021cost, cost_sensitive2022cost};
this second line of research exists because the highest UCC is often associated with misclassifying the minority class: in such cases, addressing class imbalance may also reduce TCC~\cite{imbalance2021performance}.

However, when the goal is to maximize RoI and minimize TCC, correcting class imbalance \textit{per se} may not be beneficial, especially if the minority classes are associated with misclassification costs comparable to, or lower than, those of the majority classes, and  the class distribution in the \emph{target dataset} (i.e., the real-world data where the model will be deployed) is similar to that of the \emph{development dataset} (used for training, validation, and testing).
Conceptually, the ideal approach to optimize a classifier with the final goal of maximizing RoI would be
(i) estimating the UCC for each class pair $(i, j)$, and  
(ii) minimizing the TCC computed from the UCCs and the confusion matrix.
However, this purely cost-based approach suffers from limitations:  
(i) estimating UCCs may be challenging; 
(ii) TCC depends on the dataset it is measured on, making it unsuitable for comparing classifiers across datasets; 
(iii) there is no guarantee that minimizing TCC on the development dataset will also minimize it on the target dataset; 
and (iv) in many real-world use cases UCCs depend on the specific example, challenging the assumptions underlying confusion-matrix-based evaluation approaches.

Our work provides three main contributions addressing these limitations:
(i) as an alternative to existing metrics, we introduce Weighted Accuracy (WA), an evaluation metric for binary classifiers with a straightforward interpretation as a weighted version of the well-known accuracy metric, whose maximization we prove to be equivalent to TCC minimization under example-independent UCCs;
(ii) we formalize a reweighting framework for handling class imbalance in cost-sensitive evaluation scenarios, providing an alternative to rebalancing techniques; this framework can be applied to any linear example-dependent metric, formalizes WA as a cost-based reweighted version of standard accuracy, enables consistent comparison of results obtained on different datasets, and naturally accounts for discrepancies between the class distributions in the development and target datasets;
and (iii) we provide empirical evidence of the robustness of WA by analyzing its correlation with TCC across diverse example-dependent scenarios, while also studying the validity limits of the example-independent approximation underlying WA and other confusion-matrix-based evaluation methods.

\section{Related Work}

\subsection{Confusion Matrix-Based Metrics}

The prediction output of a binary classifier is typically described in terms of a $2 \times 2$ confusion matrix, as shown in Table \ref{tab:confusion_matrix},
%
\begin{table}[ht]
\centering
\caption{Confusion matrix $\mathcal{N}$ for a binary classifier.}
\label{tab:confusion_matrix}
  \begin{tabular}{cccc}
  \toprule
   \hspace{2cm} & Predicted $+$ & Predicted $-$ & \\
  \midrule
  Actual $+$ & $\mathrm{TP}$ & $\mathrm{FN}$ & \\
   Actual $-$ & $\mathrm{FP}$ & $\mathrm{TN}$ &   \\
 \bottomrule
\end{tabular}
\end{table}
%
where $\mathrm{TP}$ ($\mathrm{TN}$) denotes the count of true positives (true negatives), $\mathrm{FP}$ ($\mathrm{FN}$) denotes the count of false positives (false negatives), while $\mathrm{P} = \mathrm{TP} + \mathrm{FN}$ ($\mathrm{N} = \mathrm{TN}+\mathrm{FP}$) is the number of positive (negative) examples, and $N_{\mathrm{tot}} = \mathrm{P}+ \mathrm{N}$ the  total number of examples.

A wide range of evaluation metrics, summarized in Table \ref{tab:metrics_confusion_matrix}, are derived from the confusion matrix, including accuracy, Jaccard's similarity index, F-measure, recall, precision, specificity, Negative Predictive Value (NPV), informedness, markedness, Matthews Correlation Coefficient (MCC), Cohen's Kappa, G-Mean, and ROC-AUC.
Despite their widespread use, each of these metrics 
has received criticism \cite{new_metrics2009measuring, new_metrics2022extending}.
To mitigate these issues, various techniques have been proposed, such as resampling strategies and cost-sensitive learning. 
However, resampling may alter the statistical properties of the original dataset, introduce redundancy, and increase computational cost~\cite{cost_sensitive1999metacost, cost_sensitive2014imbalanced, new_metrics2019metric, imbalance2021performance}.

Recent work has focused on defining confusion-matrix-based metrics to better handle class imbalance \cite{new_metrics2019metric, new_metrics2020imbalance, metrics2023metrics}; table \ref{tab:metrics_confusion_matrix} includes several such metrics; 
among these are
Class Balance Accuracy (CBA) and Imbalance Accuracy Metric (IAM) \cite{new_metrics2020imbalance},
the P4 metric \cite{new_metrics2022extending}, 
Bayesian ROC (B-ROC) curves~\cite{new_metrics2006b}.

In addition, several cost-sensitive metrics have been proposed to explicitly incorporate misclassification costs; these are also summarized in Table \ref{tab:metrics_confusion_matrix} and will be described in Section \ref{sec:related_work_cost}.

%
%
\begin{table}[ht]
\centering
\footnotesize
 \caption{Summary of confusion-matrix-based metrics for binary classifiers. Top: standard metrics; middle: recently introduced metrics; bottom: cost-sensitive metrics.}
\label{tab:metrics_confusion_matrix}
\begin{tabular}{cc}
\toprule
 Name & Formula \\
\midrule
 \multicolumn{2}{c}{\emph{Standard CM metrics}}\\
 Accuracy ($A$) & $(\mathrm{TP}+\mathrm{TN}) / N_{\mathrm{tot}}$ \\
 Recall (Sensitivity) & $\mathrm{TP} / \mathrm{P}$\\
 Precision & $\mathrm{TP} / (\mathrm{TP}+ \mathrm{FP})$\\
 Specificity & $\mathrm{TN} / \mathrm{N}$\\
 NPV & $\mathrm{TN} / (\mathrm{TN}+ \mathrm{FN})$\\
 Jaccard's similarity & $ \mathrm{TP} / (\mathrm{TP} + \mathrm{FP} + \mathrm{FN}) $ \\
 F-measure ($F_\beta$) & $(1+\beta^2)\cdot\mathrm{TP} / \left[\mathrm{TP} +\beta^2\cdot \mathrm{P} + \mathrm{FP}\right]$ \\
 Informedness & $\mathrm{TP} / \mathrm{P} -\mathrm{FP} / \mathrm{N}$ \\
 Markedness & $\mathrm{TP} / (\mathrm{TP}+\mathrm{FP})-\mathrm{FN} / (\mathrm{TN}+\mathrm{FN})$ \\
 MCC & $(\mathrm{TP} \cdot \mathrm{TN}-\mathrm{FP} \cdot \mathrm{FN}) / \sqrt{(\mathrm{TP}+\mathrm{FP})\cdot \mathrm{P}\cdot \mathrm{N}\cdot (\mathrm{TN}+ \mathrm{FN})}$\\
Cohen's kappa ($\kappa$) & $2\cdot(\mathrm{TP} \cdot \mathrm{TN} - \mathrm{FN} \cdot \mathrm{FP}) / \left[(\mathrm{TP}+\mathrm{FP})\cdot \mathrm{N}+\mathrm{P}\cdot (\mathrm{FN} +\mathrm{TN})\right]$ \\
 G-Mean & $\sqrt{\mathrm{TP} \cdot \mathrm{TN} / (\mathrm{P} \cdot \mathrm{N})}$ \\
 ROC-AUC (single param.)\footnotemark[1] & $(\mathrm{TP} / \mathrm{P} +\mathrm{TN} / \mathrm{N}) / 2$ \\
\midrule
 \multicolumn{2}{c}{\emph{Recently proposed CM metrics}}\\
 CBA & $\left[\mathrm{TP} / \max(\mathrm{P}, \mathrm{TP}+\mathrm{FP})+ \mathrm{TN} / \max(\mathrm{N}, \mathrm{TN}+\mathrm{FN})\right] / 2$\\
 IAM & $\frac{\mathrm{TP} - \max(\mathrm{FP}, \mathrm{FN})}{2 \max(\mathrm{P}, \mathrm{TP} + \mathrm{FP})} + \frac{\mathrm{TN} - \max(\mathrm{FP}, \mathrm{FN})}{2 \max(\mathrm{N}, \mathrm{TN} + \mathrm{FN})}$  \\
 P4 & $(4 \cdot \mathrm{TP} \cdot \mathrm{TN}) / \left[4 \cdot \mathrm{TP} \cdot \mathrm{TN} +(\mathrm{TP}+\mathrm{TN})\cdot(\mathrm{FP} +\mathrm{FN})\right]$ \\
  B-ROC (single param.)\footnotemark[1] & $\left[\mathrm{TP} / \mathrm{P} +\mathrm{TP} / (\mathrm{FP}+\mathrm{TP})\right] / 2$ \\
\midrule
 \multicolumn{2}{c}{\emph{Cost-sensitive metrics}} \\
 WCA\footnotemark[2] & $w \cdot \mathrm{TP} / \mathrm{P} +(1-w) \cdot \mathrm{TN} / \mathrm{N}$ \\
 WRA & $\left[4\cdot(\mathrm{TP} / \mathrm{P} -\mathrm{FP} / \mathrm{N}) \mathrm{N}\cdot C_{\mathrm{FP}} / (\mathrm{P}\cdot C_{\mathrm{FN}})\right] / \left[1+\mathrm{N}\cdot C_{\mathrm{FP}} / (\mathrm{P}\, C_{\mathrm{FN}})\right]^2$ \\
 ACD& $\sqrt{(1-A)^2 + ({\mathrm{TCC}} / {\mathrm{TCC}}_{\mathrm{max}})^2}$ \\
 C-score& ${\mathrm{TCC}} / (\mathrm{P}\cdot C_{\mathrm{FP}})$ \\
 MSU & $1 - [{\mathrm{TCC}} - {\mathrm{TCC}}_{\mathrm{min}}] / {\mathrm{TCC}}_{\mathrm{max}}$ \\
 H& $1 - \int_0^1 \mathrm{d}c\, u(c)\, {\mathrm{TCC}}(c) / \int_0^1 \mathrm{d}c\, u(c)\, {\mathrm{TCC}}_{\mathrm{max}}(c)$ \\
 \bottomrule
\end{tabular}
\footnotetext[1]{ROC-AUC and B-ROC are defined for parametric classifiers. For single-parameter classifiers, they can be computed as the area under the segments connecting the classifier point to (0,0) and (1,1) in the (TP-rate, FP-rate) or (TP-rate, False Alarm Rate) plane~\cite{metrics2011evaluation}.}
\footnotetext[2]{In WCA definition, $w$ is the sample importance, which can be identified with $C_{\mathrm{FN}} / (C_{\mathrm{FN}}+C_{\mathrm{FP}})$.}
\end{table}
%
%

\subsection{Cost-Sensitive Evaluation and Classification Costs}
\label{sec:related_work_cost}

As discussed in Section \ref{sec1}, the ideal strategy for selecting the optimal classifier in a given use case -- particularly when the goal is maximizing the RoI -- is to minimize the TCC.
In general, TCC may depend on the specific examples that are misclassified%
\footnote{For instance, in churn prediction, the cost of a false negative may vary depending on which is the misclassified customer.}
\cite{cost_sensitive2021multi} and for binary classification can be defined as
\begin{equation}\label{eq:total_cost_example_dep}
 {\mathrm{TCC}} = \sum_{a \in \mathbf{S}_+}\left(d_a^{\mathrm{FN}}\delta_{o(a),-1} +d_a^{\mathrm{TP}}\delta_{o(a),1}\right) +\sum_{a \in \mathbf{S}_-}\left(e_a^{\mathrm{TN}}\delta_{o(a),-1}+ e_a^{\mathrm{FP}}\delta_{o(a),1}\right),
\end{equation}
where $\mathbf{S}_+$ ($\mathbf{S}_-$) is the set of positive (negative) examples, $d_a^{\mathrm{FN}}$ ($d_a^{\mathrm{TP}}$) is the UCC for incorrect (correct) classification of positive example $a$, $e_a^{\mathrm{FP}}$ ($e_a^{\mathrm{TN}}$) is the UCC for incorrect (correct) classification of negative example $a$, $o(a)$ is the classifier output for example $a$, and $\delta_{jk}$ is the Kronecker delta.

In most literature, UCCs are assumed to be example-independent, i.e., $d_a^{\mathrm{FN}}=c_{\mathrm{FN}}$, $d_a^{\mathrm{TP}} = c_{\mathrm{TP}}$, $e_a^{\mathrm{FP}} = c_{\mathrm{FP}}$, and $e_a^{\mathrm{TN}}=c_{\mathrm{TN}}$. 
Under this assumption, the TCC can be expressed in terms of the confusion matrix: ${\mathrm{TCC}} = c_{\mathrm{TP}}\cdot \mathrm{TP}+ c_{\mathrm{FN}}\cdot \mathrm{FN} + c_{\mathrm{FP}}\cdot \mathrm{FP} +c_{\mathrm{TN}}\cdot \mathrm{TN}.$
To ensure economic coherence, we require $c_{\mathrm{TP}} < c_{\mathrm{FN}}$ and $c_{\mathrm{TN}} < c_{\mathrm{FP}}$~\cite{cost_sensitive2001foundations} 
and thus we can simplify the notation, introducing the shifted UCCs $C_{\mathrm{FN}} = c_{\mathrm{FN}} - c_{\mathrm{TP}}$ and $C_{\mathrm{FP}} = c_{\mathrm{FP}} - c_{\mathrm{TN}}$, obtaining
\begin{equation}
\label{eq:total_cost}
{\mathrm{TCC}} = C_{\mathrm{FN}}\cdot \mathrm{FN}+C_{\mathrm{FP}}\cdot \mathrm{FP} + {\mathrm{TCC}}_{\mathrm{min}},
\end{equation}
where ${\mathrm{TCC}}_{\mathrm{min}} = c_{\mathrm{TN}}\cdot \mathrm{N} +c_{\mathrm{TP}}\cdot \mathrm{P}$ is the minimum achievable cost, corresponding to perfect classification. 

Several studies have proposed evaluation metrics explicitly depending on TCC, as summarized in Table \ref{tab:metrics_confusion_matrix}, including
Accuracy-Cost-Distance (ACD) \cite{new_metrics2016cost},
Weighted Relative Accuracy (WRA) \cite{metrics1999rule},
Weighted Classification Accuracy (WCA) \cite{new_metrics2019learning}, and 
C-score \cite{new_metrics2024f_1}, a rescaled and dimensionless version of the TCC that correctly captures the TCC ranking, albeit its unbounded nature may pose challenges for interpretability and cross-dataset comparability.
In the context of decision theory, \cite{new_metrics2006mean} defined Mean Subjective Utility (MSU) as a normalized transformation of the utility matrix;
when adapted to the cost formalism adopted in this work, MSU becomes proportional to TCC and ranges from 1 (minimum TCC) to the lower bound $-{\mathrm{TCC}}_{\mathrm{min}}/{\mathrm{TCC}}_{\mathrm{max}}$ (maximum TCC), which depends on the dataset and cost matrix.

Finally, \cite{new_metrics2009measuring} proposed a cost-aware alternative to ROC-AUC for parametric classifiers, expressing TCC in terms of the UCC ratio $c = C_{\mathrm{FP}} / (C_{\mathrm{FP}}+C_{\mathrm{FN}})$, distributed according to a probability density function $u(c)$. The resulting metric is defined in Table \ref{tab:metrics_confusion_matrix},
where ${\mathrm{TCC}}(c) = b[c\, \mathrm{FP} + (1-c) \mathrm{FN}]$ and ${\mathrm{TCC}}_{\mathrm{max}}(c) =b[c \, \mathrm{N} + (1-c) \mathrm{P}]$, with $b = C_{\mathrm{FP}}+C_{\mathrm{FN}}$.
The H measure can be interpreted as a generalization of MSU, extending it from the edge case where $c$ is known, to scenarios where it is uncertain. In the limit case in which no prior information about $c$ is available, \cite{new_metrics2009measuring} propose modeling $u(c)$ using a Beta distribution with parameters $\alpha = \beta = 2$.

Among the existing evaluation metrics, C-score, MSU, and H stand out for their linear relation with TCC, allowing for model ranking coherent with TCC. However, each has limitations. C-score, while dimensionless,
lacks an upper bound, which can complicate comparisons
across different datasets, as its magnitude depends on dataset-specific properties, such as the number of examples and the proportion of actual positives. MSU and H are normalized within the $[0,1]$ interval only when ${\mathrm{TCC}}_{\mathrm{min}} = 0$; if this condition is not met, the absence of a consistent lower bound may affect
interpretability and comparability.

As a better alternative to  C-score, MSU, and H, in the next section we propose  a weighted version of the well-known accuracy metric, normalized between 0 and 1.
In the case of example-independent Unit Classification Costs (UCCs), this metric not only ranks classifiers consistently with the Total Classification Cost (TCC), but also enables performance comparison across different datasets and supports TCC minimization on the target dataset, provided that the ratio of positive examples (or the prior probability of a positive outcome) can be estimated.

\section{Weighted Accuracy and Cost-Sensitive Reweighting}
\label{sec:WA_and_framework}

\subsection{Weighted Accuracy}
\label{sec:WA}

Accuracy, as defined in Table \ref{tab:metrics_confusion_matrix}, is a standard performance indicator, widely used to evaluate classifiers due to its simplicity: it measures the proportion of correct predictions over the total number of examples. However, it assumes equal importance for all types of classification outcomes, deviating from TCC in the presence of unequal misclassification costs, and it is often considered inappropriate in imbalanced scenarios.

To address these limitations, we define a \textit{Weighted Accuracy} (WA), which assigns different importance to positive and negative examples
\begin{equation}
\label{eq:w_accuracy_binary}
    \mathrm{WA} = \frac{w \, \mathrm{TP} + (1-w) \mathrm{TN}}{w \, \mathrm{P} + (1-w) \mathrm{N}},
\end{equation}
where $w$ is the normalized weight assigned to actual positives (true positives and false negatives), and $1-w$ is the weight assigned to actual negatives (false positives and true negatives).
If we choose the weight $w$ as the UCC ratio, i.e.,
\begin{equation}\label{eq:w=rC}
 w=r_C,
\end{equation}
with
\begin{equation}
\label{eq:weight_from_cost}
    r_C =\frac{C_{\mathrm{FN}}}{C_{\mathrm{FN}} + C_{\mathrm{FP}}},
\end{equation}
then WA becomes linearly related to the TCC defined in Eq.~\ref{eq:total_cost}; indeed, substituting Eqs.~\ref{eq:w=rC} and \ref{eq:weight_from_cost} into Eq.~\ref{eq:w_accuracy_binary}, we obtain 
\begin{equation}\label{eq:WA_propto_TCC}
 \mathrm{WA} = 1 - \frac{{\mathrm{TCC}}-{\mathrm{TCC}}_{\mathrm{min}}}{{\mathrm{TCC}}_{\mathrm{max}}-{\mathrm{TCC}}_{\mathrm{min}}},
\end{equation}
where ${\mathrm{TCC}}_{\mathrm{max}} = C_{\mathrm{FN}} \mathrm{P}+ C_{\mathrm{FP}} \mathrm{N}$ is the maximum possible classification cost.
This expression shows that WA is analogous to the C-score, MSU, and H metrics discussed in Section \ref{sec:related_work_cost}, while providing a more intuitive interpretation and a more consistent normalization scheme. It also demonstrates that Eq.~\ref{eq:w_accuracy_binary} offers a more principled way to introduce cost-based weighting into accuracy than, for example, WCA (see Table \ref{tab:metrics_confusion_matrix}).

 Consider a scenario in which the UCC of a false negative is nine times larger than that associated with a false positive ($r_C = 0.9$), and assume that positive examples constitute $20\%$ of the dataset.
 Let $\mathcal{M}_1$ denote a classifier that always predicts the negative label, and $\mathcal{M}_2$ a classifier characterized by $\mathrm{TP}=15$, $\mathrm{FN}=5$, $\mathrm{TN}=50$, and $\mathrm{FP}=30$; the corresponding accuracies are $80\%$ and $65\%$, respectively, while $\mathrm{WA}(\mathcal{M}_1)\simeq30\% $ and $\mathrm{WA}(\mathcal{M}_2)\simeq 71\% $.

 Importantly, the failure of accuracy in this example does not arise because predicting only the majority class should \emph{per se} be penalized. Rather, it results from the strong asymmetry in the UCCs, which makes false negatives substantially more costly than false positives; indeed, when UCCs are balanced ($r_C=0.5$), WA coincides with accuracy for any classification outcome.

For example-independent UCCs, WA with the weight in Eq.~\ref{eq:weight_from_cost} ranks confusion matrices in the exact reverse order of TCC. Therefore, maximizing WA is equivalent to minimizing TCC and, by extension, maximizing RoI. 
By construction, WA is normalized to the interval $[0, 1]$ and is independent of the test set size; this makes it suitable for comparing models evaluated on different datasets, as long as the class distribution (i.e., $\mathrm{P} / N_{\mathrm{tot}}$) remains the same.
In Sections \ref{sec:handling_class_imbalance} and \ref{sec:cost_based_reweighting}, 
we show how WA can be straightforwardly adapted  to compare results across datasets with different class distributions by appropriately adjusting the weight $w$.

\subsection{Expected Weighted Accuracy}\label{sec:EWA}

When the value of $w$ cannot be precisely determined, a probabilistic approach may be adopted: we introduce a probability density function $u(w)$ over the weight $w$, allowing us to define the \emph{Expected Weighted Accuracy} (EWA)
\begin{equation}
\label{eq:expected_weighted_accuracy}
    \mathrm{EWA} = \int_0^1 \mathrm{d}w\, \mathrm{WA}(w)\, u(w).
\end{equation}
Although this formulation shares similarities with the H-measure introduced in~\cite{new_metrics2012novel, new_metrics2009measuring}, it is not equivalent, since in the H-measure, the integrals of ${\mathrm{TCC}}(c)$ and ${\mathrm{TCC}}_{\mathrm{max}}(c)$ are computed separately, as shown in Table \ref{tab:metrics_confusion_matrix}.

Depending on the available information about the cost distribution, $u(w)$ can be modeled using a Beta distribution (as discussed in Section \ref{sec:related_work_cost}), or constructed as a custom distribution over a plausible range of $w$ values. If the support of $u(w)$ is narrow -- e.g., $u(w) > 0$ only within a small interval -- then EWA can be approximated by WA. 
For instance, for a narrow interval $[\bar{w} - \delta, \bar{w}+\delta]$, we have
$ \mathrm{EWA} = \mathrm{WA}(\bar{w}) + O(\delta^2)$,
which implies that EWA can be well approximated by evaluating WA at the midpoint $\bar{w}$ of the interval.

\subsection{Handling Class Imbalance through Reweighting}
\label{sec:handling_class_imbalance}

We propose a reweighting approach for handling class imbalance in classifier evaluation that avoids resampling, thereby preserving the original dataset and its statistical properties.

Consider a generic performance metric that can be expressed as a weighted average:
\begin{equation}
\label{eq:kpi_average}
    K = \frac{\sum_a v_a k_a}{\sum_a v_a},
\end{equation}
where $k_a$ is the contribution of example $a$, $v_a$ its associated weight, and the sum runs over all examples in the original set $\mathbf{S}$. 

Accuracy and WA are special cases of this formulation, with $k_a = 1$ ($0$) if example $a$ is correctly (incorrectly) classified.  
For accuracy, all examples contribute uniformly:
\begin{equation}\label{eq:v_a_accuracy}
 v_a(\mathrm{accuracy})=\frac{1}{N_{\mathrm{tot}}},
\end{equation}
while WA (defined in Eq. \ref{eq:w_accuracy_binary}) introduces class-dependent weights:
\begin{equation}\label{eq:v_a_WA}
  v_a(\mathrm{WA}) = 
\begin{cases}
\frac{w}{w \mathrm{P} + (1-w) \mathrm{N}} & \text{if } a \in \mathbf{S}_+ \\
\frac{1-w}{w \mathrm{P} + (1-w) \mathrm{N}} & \text{if } a \in \mathbf{S}_-
\end{cases}.
\end{equation}

Assume that we wish to estimate the value of $K$ on a balanced version of $\mathbf{S}$ containing the same number of positive and negative examples%
\footnote{The generalization to arbitrary class ratios is discussed in Section \ref{sec:target_set_balancing}.}.
Rather than explicitly constructing such a dataset through rebalancing techniques, we can equivalently
rescale each weight $v_a$ according to the relative frequency of its class; after normalization, this procedure yields
\begin{equation}
\label{eq:weight_correct_imbalance_kpi}
v_a^{\mathrm{balanced}} = 
\begin{cases}
\frac{\mathrm{N}\,v_a}{\mathrm{N}\,V_+ +\mathrm{P}\,V_- } & \text{if } a \in \mathbf{S}_+ \\
\frac{\mathrm{P}\,v_a}{\mathrm{N}\,V_+ +\mathrm{P}\,V_- } & \text{if } a \in \mathbf{S}_-
\end{cases},
\end{equation}
where $V_+=\sum_{a\in\mathbf{S}_+}v_a$ and $V_-=\sum_{a\in\mathbf{S}_-}v_a$.

This formulation is consistent with previous findings \cite{cost_sensitive2006influence, cost_sensitive2008cost, imbalance2018curse, imbalance2020streams, cost_sensitive2021cost}. Although the focus of this work is on classifier evaluation, the proposed reweighting framework applies to any metric of the form in Eq.~\ref{eq:kpi_average}, including loss functions (see Section \ref{sec:framework_for_model_training}). 

In particular, for accuracy, the rebalanced weights become 
\begin{equation}
\label{eq:weight_rebalanced_accuracy}
v_a^{\mathrm{balanced}}(\mathrm{accuracy}) = 
\begin{cases}
\frac{1}{2\mathrm{P}} & \text{if } a \in \mathbf{S}_+ \\
\frac{1}{2\mathrm{N}} & \text{if } a \in \mathbf{S}_-
\end{cases},
\end{equation}
which corresponds to evaluating accuracy on a perfectly balanced dataset.

\subsection{Cost-Based Reweighting}
\label{sec:cost_based_reweighting}

We now extend the metric in Eq.~\ref{eq:kpi_average} to the case of unequal, example-independent UCCs $C_{\mathrm{FN}}$ and $C_{\mathrm{FP}}$;
to account for asymmetric UCCs, we introduce a class-dependent reweighting scheme in which positive and negative examples are weighted proportionally to $C_{\mathrm{FN}}$ and $C_{\mathrm{FP}}$, respectively. After normalization, the transformed weights become
\begin{equation}
\label{eq:w_target_kpi_unequal_UCCs_no_target}
v_a^{\mathrm{UCC}} = 
\begin{cases}
\frac{r_C v_a}{r_C V_++(1-r_C)V_-}& \text{if } a \in \mathbf{S}_+ \\
\frac{(1-r_C) v_a}{r_C V_++(1-r_C)V_-}& \text{if } a \in \mathbf{S}_-
\end{cases}.
\end{equation}

Applying this cost-based reweighting scheme to accuracy yields 
\begin{equation}
\label{eq:weight_UCC_accuracy}
v_a^{\mathrm{UCC}}(\mathrm{accuracy}) = 
\begin{cases}
\frac{r_C}{r_C\mathrm{P}+(1-r_C)\mathrm{N}} & \text{if } a \in \mathbf{S}_+ \\
\frac{1-r_C}{r_C\mathrm{P}+(1-r_C)\mathrm{N}} & \text{if } a \in \mathbf{S}_-
\end{cases}.
\end{equation}
Equation~\ref{eq:weight_UCC_accuracy} coincides with Eq.~\ref{eq:v_a_WA} when the weight $w$ is defined as in Eq.~\ref{eq:w=rC}; therefore, WA can be interpreted as a cost-reweighted version of standard accuracy according to the proposed framework.

As discussed in Section \ref{sec:WA}, maximizing WA is equivalent to minimizing TCC in the presence of example-independent UCCs (see Eq.~\ref{eq:WA_propto_TCC}). 
Comparing the balancing-based reweighting in Eq.~\ref{eq:weight_rebalanced_accuracy} with the cost-based reweighting in Eq.~\ref{eq:weight_UCC_accuracy}, we conclude that maximizing accuracy on a rebalanced dataset is equivalent to minimizing TCC only when $1/(2\mathrm{P}) \simeq r_C/\left[r_C\mathrm{P} + (1-r_C)\mathrm{N}\right]$, i.e., when 
\begin{equation}\label{eq:rescaling_regime}
 r_C \simeq \frac{N}{N_{\mathrm{tot}}}.
\end{equation}
This condition is implicitly assumed by standard rebalancing techniques and by several metrics designed to correct class imbalance, yet it does not hold in general; when Eq.~\ref{eq:rescaling_regime} is violated, such approaches may produce substantially misleading evaluations.

\subsection{Target Dataset Reweighting}
\label{sec:target_set_balancing}

The weighting scheme defined in Eq.~\ref{eq:weight_correct_imbalance_kpi} aims to correct class imbalance, or equivalently estimating the metric after positive and negative classes are balanced to the same fraction of examples.
However, for cost-sensitive evaluation, the relevant quantity is the  class distribution in the \textit{target dataset}, i.e., the dataset on which the classifier will be deployed and decisions will be made. This could correspond to a production environment in business or a patient population in healthcare.
Let us then distinguish between:
(\emph{i}) the \textit{development dataset}, with $N_{\mathrm{tot}}$ examples and ratio of positive examples $r_+ = \mathrm{P} / N_{\mathrm{tot}}$, used for training, validation, and testing and
(\emph{ii}) the \textit{target dataset}, with $\mathrm{N}_{\mathrm{tot}}^{\mathrm{t}}$ examples and ratio of positive examples $r_+^{\mathrm{t}} = \mathrm{P}_{\mathrm{t}} / N_{\mathrm{tot}}^{\mathrm{t}}$, which may differ from $r_+$.
If the target dataset is not directly accessible, $r_+^{\mathrm{t}}$ can be interpreted as a \textit{base rate}, i.e., the prior probability of a positive outcome~\cite{new_metrics2024f_1}.

To estimate a metric on the target dataset using the development dataset, we generalize Eq.~\ref{eq:weight_correct_imbalance_kpi} as:
\begin{equation}
\label{eq:w_target_kpi}
v_a^{\mathrm{t}} = 
\begin{cases}
\frac{R_+^{\mathrm{t}}v_a }{R_+^{\mathrm{t}} V_+ + R_-^{\mathrm{t}} V_-} & \text{if } a \in \mathbf{S}_+ \\
\frac{R_-^{\mathrm{t}}v_a }{R_+^{\mathrm{t}} V_+ + R_-^{\mathrm{t}} V_-} & \text{if } a \in \mathbf{S}_-
\end{cases},
\end{equation}
where $R_+^{\mathrm{t}}=r_+^{\mathrm{t}}/r_+$ and $R_-^{\mathrm{t}}=(1-r_+^{\mathrm{t}})/(1-r_+)$.
As expected, Eq.~\ref{eq:w_target_kpi} reduces to Eq.~\ref{eq:weight_correct_imbalance_kpi}
if the target dataset is perfectly balanced, i.e., if $r_+^{\mathrm{t}} = 0.5$.

This rescaling also enables consistent comparison of metrics calculated across datasets with different class distributions by mapping them to a common target distribution.

To further extend Eq.~\ref{eq:w_target_kpi} to the case of unequal, example-independent UCCs $C_{\mathrm{FN}}$ and $C_{\mathrm{FP}}$, 
we combine target-distribution reweighting with the cost-based reweighting scheme introduced in Eq.~\ref{eq:w_target_kpi_unequal_UCCs_no_target}, obtaining
\begin{equation}
\label{eq:w_target_kpi_unequal_UCCs}
v_a^{\mathrm{t, UCC}} = 
\begin{cases}
\frac{R_+^{\mathrm{t}}r_Cv_a }{R_+^{\mathrm{t}}r_C V_+ + R_-^{\mathrm{t}}(1-r_C) V_-} & \text{if } a \in \mathbf{S}_+ \\
\frac{R_-^{\mathrm{t}}(1-r_C)v_a }{R_+^{\mathrm{t}}r_C V_+ + R_-^{\mathrm{t}}(1-r_C) V_-} & \text{if } a \in \mathbf{S}_-
\end{cases},
\end{equation}
where $r_C$ is defined in Eq. \ref{eq:weight_from_cost}.

Equation \ref{eq:w_target_kpi_unequal_UCCs} provides a unified framework to estimate metrics of the form in Eq.~\ref{eq:kpi_average} while jointly accounting for unequal UCCs and differences between the class distributions of the development and target datasets.
If misclassification costs are equal ($r_C = 1/2$), it reduces to Eq.~\ref{eq:w_target_kpi}; on the other hand, if the class proportion in target and development datasets coincide ($r_+^{\mathrm{t}} = r_+$), it reduces to Eq.~\ref{eq:w_target_kpi_unequal_UCCs_no_target}.

Applying Eq.~\ref{eq:w_target_kpi_unequal_UCCs} to the uniform weights of accuracy (Eq.~\ref{eq:v_a_accuracy}) 
yields a WA formulation (Eq.~\ref{eq:v_a_WA}) with class weight
\begin{equation}
\label{eq:w_target_accuracy}
    w_{\mathrm{t}} = \frac{R_+^{\mathrm{t}}r_C}{R_+^{\mathrm{t}}r_C + R_-^{\mathrm{t}}(1-r_C)}.
\end{equation}

Equation \ref{eq:w_target_accuracy} determines the WA weight provided that the UCC ratio $r_C$ and the ratio of positives $r_+^{\mathrm{t}}$ in the target dataset can be estimated. In Appendix \ref{app:weight_estimation}, we describe a procedure to estimate the weight $w$ in the presence of uncertainty in $r_C$.
If substantial uncertainty on $w$ remains after this estimation procedure, the probabilistic approach introduced in Section \ref{sec:EWA} can be adopted.

Finally, Eq.~\ref{eq:w_target_accuracy} shows that a weighting scheme (or, alternatively, class rebalancing) is required for accuracy unless
\begin{equation}\label{eq:accuracy_validity_target}
r_+\simeq\frac{r_+^{\mathrm{t}}r_C}{1-r_+^{\mathrm{t}}-r_C +2r_+^{\mathrm{t}}r_C}
\end{equation}
which is therefore the condition under which standard accuracy is equivalent to TCC. 
If the target dataset has the same positive rate as the development dataset ($r_+^{\mathrm{t}} \simeq r_+$), Eq. \ref{eq:accuracy_validity_target} reduces to the equal-UCCs condition ($r_C\simeq 1/2$).

\subsection{Example-Dependent Unit Classification Costs}
\label{sec:example_dependent_TCC}

Although most confusion-matrix-based literature assumes that UCCs are example-independent, in many real-world use cases they actually depend on the individual example $a$~\cite{cost_sensitive2019example, cost_sensitive2021multi}.
In such cases, evaluating the TCC (see Eq.~\ref{eq:total_cost_example_dep}) requires specifying the set $\mathbf{S}_{\mathrm{FN}}$ ($\mathbf{S}_{\mathrm{FP}}$) of misclassified positive (negative) examples, rather than simply their count $\mathrm{FN}$ ($\mathrm{FP}$). This implies that the same confusion matrix may correspond to different TCC values depending on which examples are misclassified.

We can extend the shifted UCCs defined in Section \ref{sec:related_work_cost} to the example-dependent case as $E_a^{\mathrm{FP}} = e_a^{\mathrm{FP}}-e_a^{\mathrm{TN}} $ and $D_a^{\mathrm{FN}} = d_a^{\mathrm{FN}}-d_a^{\mathrm{TP}}$; this allows us to rewrite Eq.~\ref{eq:total_cost_example_dep} as
$ \mathrm{TCC} = \sum_{a \in\mathbf{S}_{\mathrm{FN}}} D_a^{\mathrm{FN}}+\sum_{a \in \mathbf{S}_{\mathrm{FP}}} E_a^{\mathrm{FP}} +\mathrm{TCC}_{\mathrm{min}}$, 
where $\mathrm{TCC}_{\mathrm{min}} = \sum_{a \in \mathbf{S}_+} d_a^{\mathrm{TP}} + \sum_{a \in \mathbf{S}_-} e_a^{\mathrm{TN}}$ is the minimum achievable cost, corresponding to perfect classification.
If we
express the UCCs as $D_a^{\mathrm{FN}} = C_{\mathrm{FN}} + \delta_a$ and $E_a^{\mathrm{FP}} = C_{\mathrm{FP}} +\epsilon_a$ -- where $C_{\mathrm{FN}}$ ($C_{\mathrm{FP}}$) is the average cost over the positives (negatives) of the target dataset%
\footnote{If these averages are not known for the target dataset, they may be estimated from the development dataset.}%
, and $\delta_a$ and $\epsilon_a$ represent example-dependent deviations -- we obtain
\begin{equation}
\label{eq:total_cost_example_dep_3}
 \mathrm{TCC} = C_{\mathrm{FN}}\cdot\mathrm{FN} + C_{\mathrm{FP}} \cdot \mathrm{FP}+\mathrm{TCC}_{\mathrm{min}} +\sum_{a \in \mathbf{S}_{\mathrm{FN}}} \delta_a  + \sum_{a\in\mathbf{S}_{\mathrm{FP}}} \epsilon_a. 
\end{equation}

Equation \ref{eq:total_cost_example_dep_3} extends Eq.~\ref{eq:total_cost} by adding terms that depend on the specific misclassified examples;
Eq.~\ref{eq:total_cost} is exact for a given classification outcome only if the averages $C_{\mathrm{FN}}$  and $C_{\mathrm{FP}}$ are computed on the subset of  misclassified positives and misclassified negatives, respectively.
Since UCCs are typically estimated \emph{a priori}, without knowing which examples will be misclassified, deviations between Eq.~\ref{eq:total_cost_example_dep_3} and Eq.~\ref{eq:total_cost} are therefore expected. For instance, 
in the limiting case where a subset of  \emph{massive examples} accounts for most of the TCC, the counts $\mathrm{FN}$ and $\mathrm{FP}$ become largely uninformative for determining the TCC, and the dominant factor is whether these massive examples are correctly classified. 
In this regime, systematically analyzed in Section \ref{sec:extreme_statistics},  the fluctuations terms $\delta_a$ and $\epsilon_a$ become decisive.

In many practical situations, however, the terms $\sum_a\delta_a$ and $\sum_a \epsilon_a$ are not prominent, even when the single fluctuations $\delta_a$ or $\epsilon_a$ are comparable to or larger than the averages $C_{\mathrm{FN}}$ and $C_{\mathrm{FP}}$. In Section \ref{sec:metrics_comparison_with_TCC} we illustrate this behavior through two representative use cases, covering all possible regimes in terms of $r_+$ and $r_C$.

\section{Empirical Evaluation against Total Classification Cost}
\label{sec:metrics_comparison_with_TCC}

Here, we compare WA, EWA (introduced in Section \ref{sec:EWA}), and the metrics described in Table \ref{tab:metrics_confusion_matrix}, using TCC as a reference in two example-dependent scenarios.

The TCC is computed according to Eq.~\ref{eq:total_cost_example_dep_3}, while the cost-sensitive metrics WCA, WRA, ACD, WA, EWA are calculated using average UCCs.
The H metric is computed using a Beta distribution with $\alpha = \beta = 2$ for the UCC ratio, as recommended by~\cite{new_metrics2009measuring} when no prior information about the cost distribution is available.
H informed and EWA are computed using a Beta distribution whose mean and variance match those of the empirical $u(c)$ distribution%
\footnote{To approximate the empirical distribution $u(c)$ with a Beta distribution $f(c;\alpha, \beta)$, we set $\alpha = \bar c^2(1-\bar c)/\sigma^2_c -\bar c$ and $\beta=\alpha(1-\bar c)/\bar c$. With this choice, $u$ and $f$ share the same mean $\bar c$ and variance $\sigma^2_c$.}.
MSU and C-score are not explicitly computed, as they rank classification outcomes identically to WA, differing only in their normalization schemes.

To quantify the similarity between each metric and TCC, we identify a sample of classification outcomes and rank them according to the metric and TCC, respectively; hence we compute the correlation between the two rankings, using:
\begin{itemize}
    \item the standard Spearman coefficient~\cite{spearman1961proof}, which corresponds to the Pearson correlation between rank variables;
    \item the weighted Spearman coefficient~\cite{new_metrics2020top,new_metrics2025standardization}, which assigns greater importance to the correct ordering of top-ranked outcomes, i.e., those with low TCC or high metric values, which are typically the focus during model selection and validation; for this coefficient we use an additive weighting scheme and $n_0 = 2$.
\end{itemize}

\subsection{Experimental Setup}
\label{sec:data_algorithm}

Real-world classification problems vary substantially in terms of both the target dataset, particularly its imbalance ratio, and the UCCs. Two key quantities characterizing each scenario are the ratio of positive examples $r_+$ and the UCC ratio $r_C$, both ranging in $[0,1]$. 
To systematically explore the full range of possible $(r_C, r_+)$ scenarios, we construct a discrete two-dimensional grid over the domain $[0,1] \times [0,1]$.
For each pair $(r_C, r_+)$ in the grid, we generate 100 samples by randomly assigning the positive label (churn) to a subset of $\mathrm{P}={\rm round}(r_+ N_{\mathrm{tot}})$ examples in each sample.
For each sample, we generate $N_{\mathrm{tot}} +1$ classification outcomes by iterating over all possible numbers $\mathrm{P}_{\mathrm{pred}}$ of predicted positives and assigning the $\mathrm{P}_{\mathrm{pred}}$ predicted positive labels to a random subset of examples in the sample.

For each metric $Y$, these outcomes are ranked according to both TCC and $Y$, generating two rankings that are compared using the correlation coefficients described above.
The correlations obtained for the 100 samples are then averaged, yielding a mean correlation value for each metric $Y$ and each point in the ($r_+, r_C$) grid.

The code used to run the experiments and generate the results described in this work is available at \url{https://github.com/plombardML/weighted-accuracy}.
Experiments were conducted on a Windows machine with 16 GB RAM and an Intel(R) Core(TM) i5-102100U CPU @ 1.60 GHz 2.10 GHz, using Python, the scipy library, and the implementation 
in \cite{new_metrics2025standardization} for computing the weighted Spearman coefficient.

\subsubsection{Use Cases}\label{sec:use_cases}

To work with data that are as realistic and relatable as possible, we consider two well-known use cases characterized by example-dependent costs; for each use case, we tune a cost parameter so that the resulting UCC ratio $r_C$ matches the values specified by the grid described above (see Appendix \ref{app:use_cases_details}).

\begin{enumerate}
  \item The first use case is \emph{churn prediction}, where the model predicts whether a customer will churn within a forthcoming time window. 
   The cost of a false positive includes the time and effort of the commercial team to contact the customer and implement a retention measure as well as the cost of that measure. Conversely, the cost of a false negative depends on the revenue the company would have earned had the customer not churned; this depends on the customer's revenue and on the probability of effectiveness of the retention measure. Customer revenues are sampled uniformly and without repetition from the monthly charges in the \emph{Telco Customer Churn} dataset from Kaggle%
\footnote{Dataset available at \url{https://www.kaggle.com/datasets/blastchar/telco-customer-churn}\label{foot:churn_dataset}},
while the remaining parameters are estimated as described in Appendix \ref{app:churn}.

  \item The second use case is \emph{credit scoring}, where the model predicts whether a customer will default on a contracted financial obligation.
  The cost of a false negative is proportional to the customer's debt (credit line). In contrast, the cost of a false positive reflects the loss of profit from rejecting a customer who would not default; since, in case of rejection, the institution typically lends to an alternative customer, this cost depends on the expected average profit and risk of a representative customer.
  The customers' monthly income and debt ratio are sampled uniformly and without repetition from the dataset of the 2011 Kaggle competition \emph{Give Me Some Credit}%
\footnote{Dataset available at \url{https://www.kaggle.com/c/GiveMeSomeCredit}}%
, while the remaining parameters are estimated as described in Appendix \ref{app:credit}.
 \end{enumerate}

In both use cases, the average UCCs and their fluctuations strongly depend on the pair $(r_C, r_+)$.
Comparing the standard deviations $\sigma_{\delta}$ and $\sigma_{\epsilon}$ of $\delta_a$ and $\epsilon_a$ with the corresponding average costs $C_{\mathrm{FN}}$  and $C_{\mathrm{FP}}$, we observe substantial variability.
For churn prediction, the ratio $\sigma_{\delta} / C_{\mathrm{FN}}$ ranges from 0.03 to 47 (median 0.79). 
For credit scoring, it ranges from 0.0002 to 51 (median 0.16), while $\sigma_{\epsilon} / C_{\mathrm{FP}}$ ranges from 0.005 to 2800 (median 2.38). 
In many $(r_C, r_+)$ scenarios the fluctuations are therefore substantial, and a close agreement between Eq.~\ref{eq:total_cost} and Eq.~\ref{eq:total_cost_example_dep_3} -- that is, between WA and the example-dependent TCC -- cannot be assumed \emph{a priori}.
In both cases the target dataset is assumed to be equivalent to the development dataset in terms of ratio of positives ($r_+=r_+^{\mathrm{t}}$), with the latter containing 200 examples ($N_{\mathrm{tot}} =200$).

\subsection{Results}
\label{sec:results}
%
\begin{figure}[ht]
    \centering
    \includegraphics[width=\columnwidth]{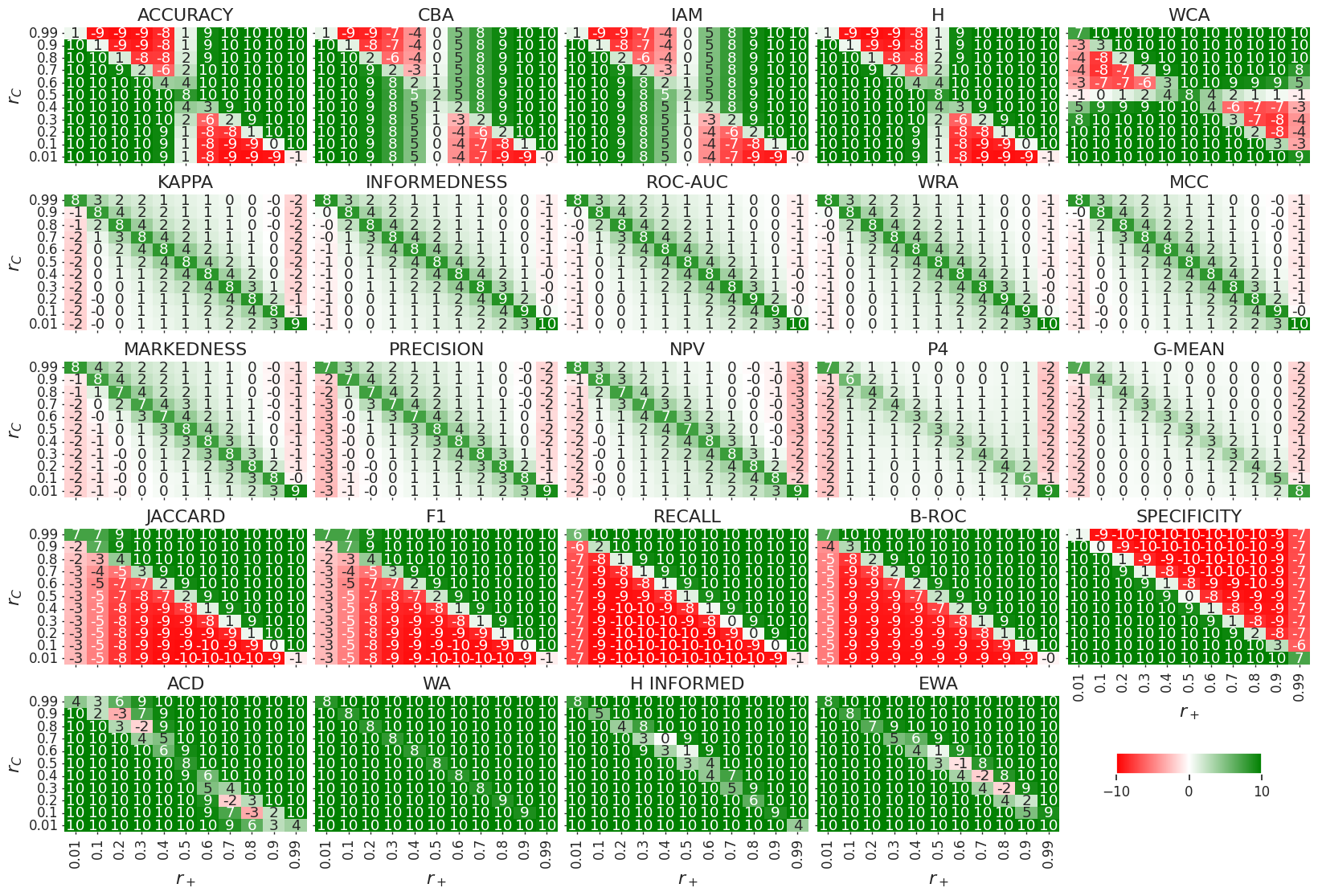}
    \caption{Heatmaps showing {\bf standard Spearman} correlation between TCC and the metrics in Table \ref{tab:metrics_confusion_matrix} and Section \ref{sec:WA} over a grid of values of $r_+$ (horizontal axis) and $r_C$ (vertical axis) for the {\bf churn prediction} use case. To improve readability, correlation coefficients are multiplied by 10 and rounded.}
    \label{fig:heatmap_churn_standard}
\end{figure}
%

%
\begin{figure}[ht]
    \centering
    \includegraphics[width=\columnwidth]{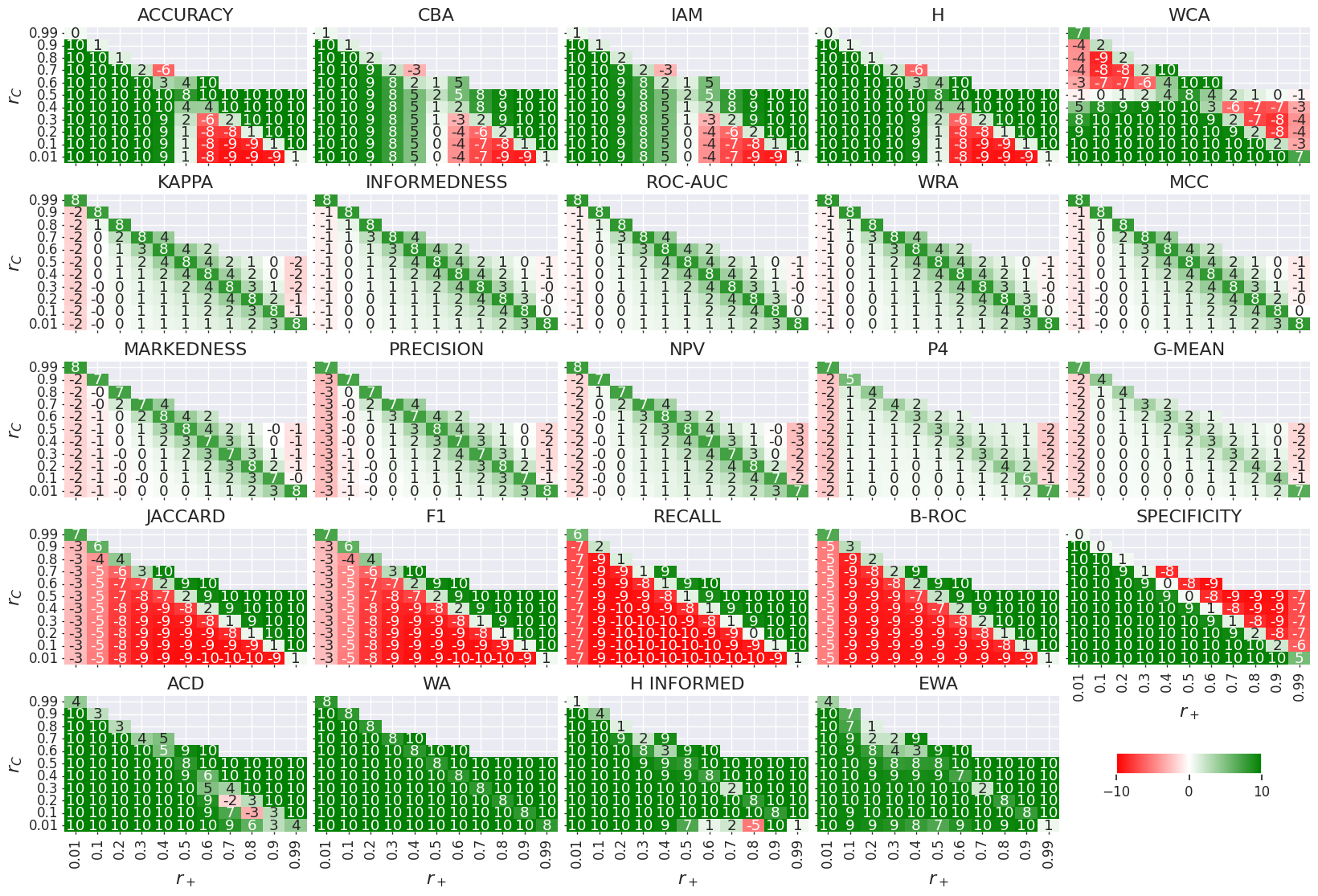}
    \caption{Heatmaps showing {\bf standard Spearman} correlation between TCC and the metrics in Table \ref{tab:metrics_confusion_matrix} and Section \ref{sec:WA} over a grid of values for $r_+$ (horizontal axis) and $r_C$ (vertical axis) for the {\bf credit scoring} use case. 
    Values with $r_C \geq 1 / (1 + r_+)$ are omitted due to consistency conditions (more details in Appendix \ref{app:credit}).
    To improve readability, correlation coefficients are multiplied by 10 and rounded.
    }
    \label{fig:heatmap_credit_standard}
\end{figure}
%

%
\begin{figure}[ht]
    \centering
    \includegraphics[width=\columnwidth]{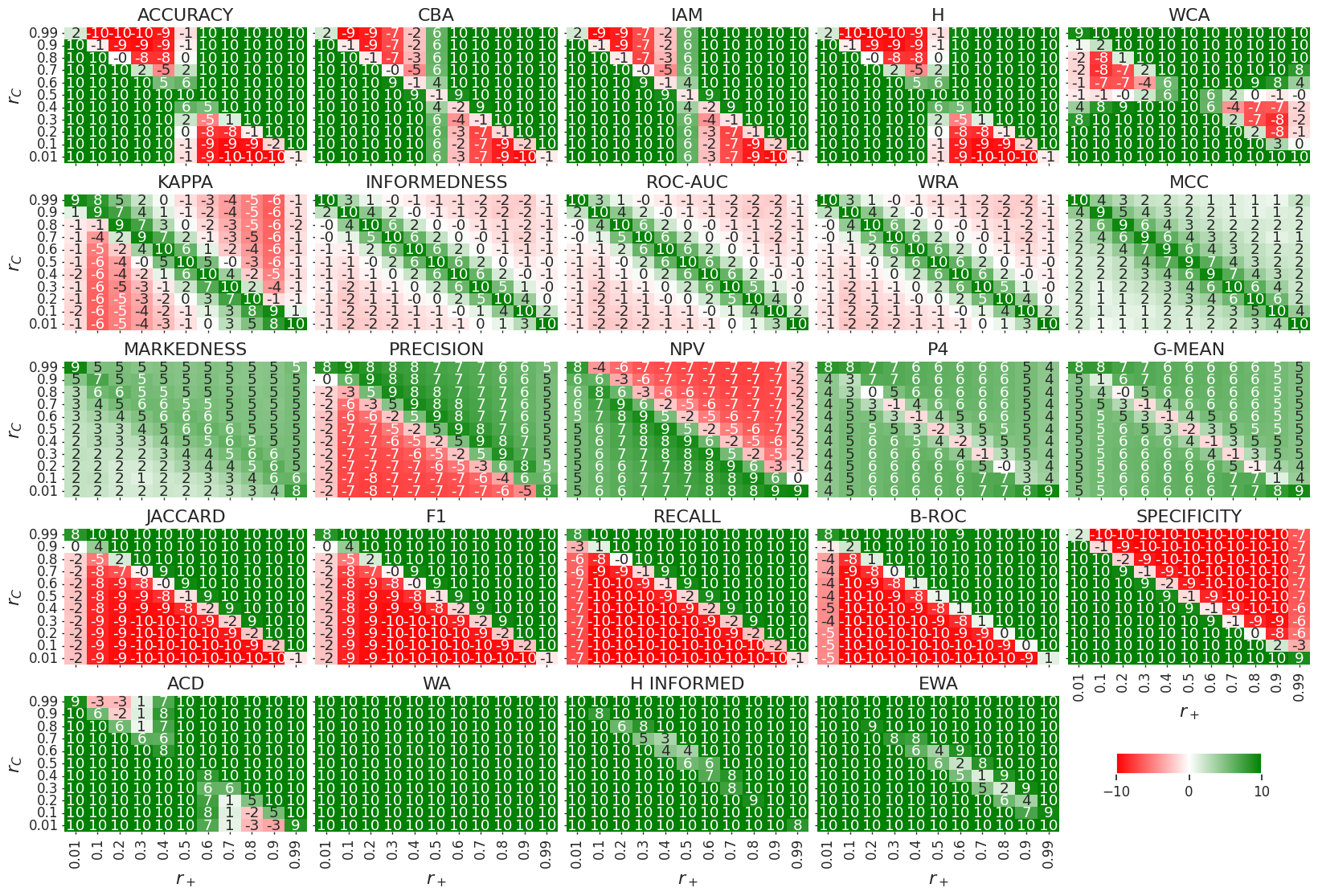}
    \caption{Heatmaps showing {\bf weighted Spearman} correlation between TCC and the metrics in Table \ref{tab:metrics_confusion_matrix} and Section \ref{sec:WA} over a grid of values of $r_+$ (horizontal axis) and $r_C$ (vertical axis) for the {\bf churn prediction} use case. To improve readability, correlation coefficients are multiplied by 10 and rounded.}
    \label{fig:heatmap_churn_weighted}
\end{figure}
%

%
\begin{figure}[ht]
    \centering
    \includegraphics[width=\columnwidth]{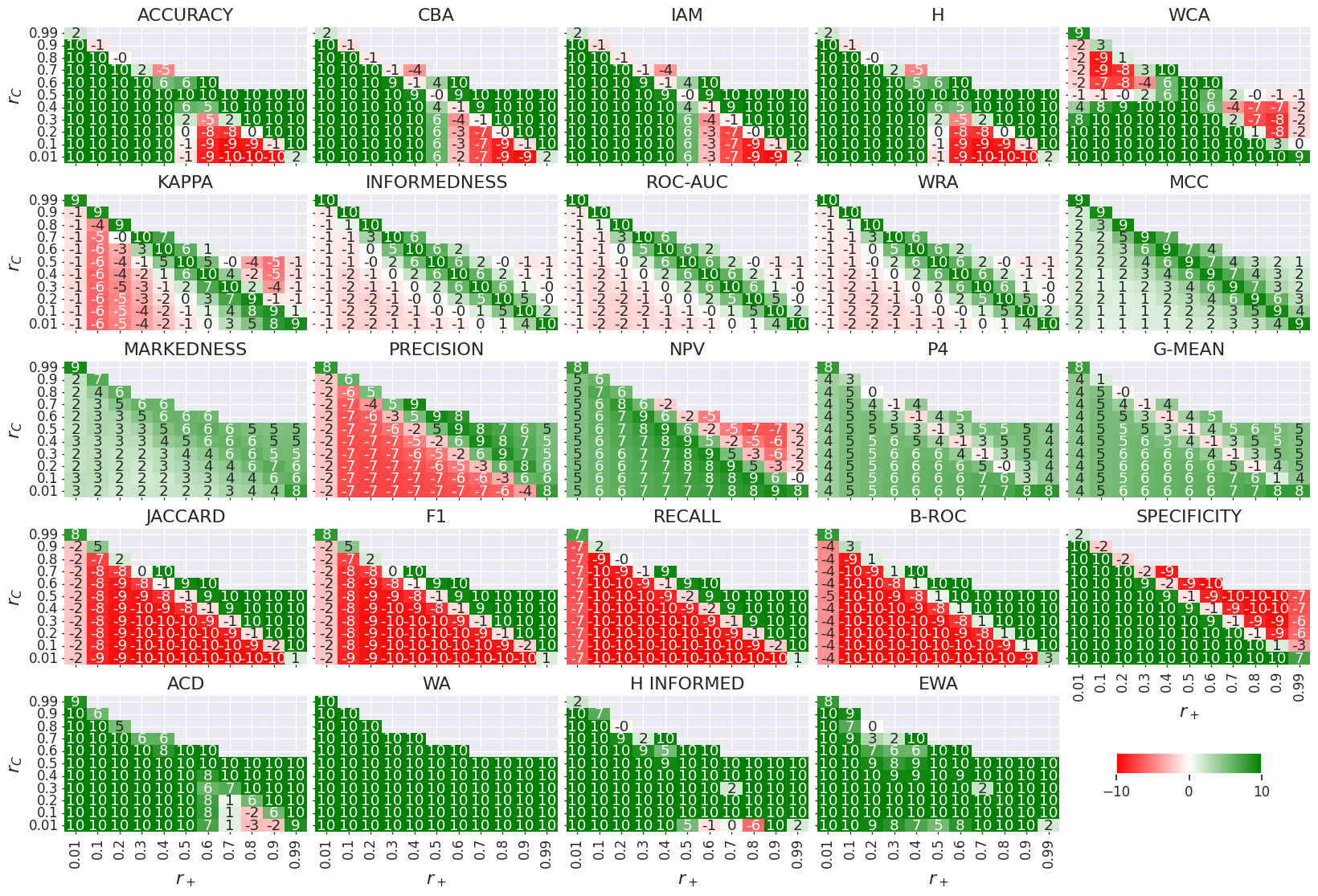}
    \caption{Heatmaps showing {\bf weighted Spearman} correlation between TCC and the metrics in Table \ref{tab:metrics_confusion_matrix} and Section \ref{sec:WA} over a grid of values for $r_+$ (horizontal axis) and $r_C$ (vertical axis) for the {\bf credit scoring} use case. 
    Values with $r_C \geq 1 / (1 + r_+)$ are omitted due to consistency conditions (more details in Appendix \ref{app:credit}).
    To improve readability, correlation coefficients are multiplied by 10 and rounded.}
    \label{fig:heatmap_credit_weighted}
\end{figure}
%

The results of this correlation analysis are shown in Figures \ref{fig:heatmap_churn_standard} and \ref{fig:heatmap_credit_standard} 
for the standard Spearman correlation in the churn prediction and credit scoring use cases, respectively,
and in Figures \ref{fig:heatmap_churn_weighted} and \ref{fig:heatmap_credit_weighted} for the corresponding weighted Spearman correlation.
Several metrics -- namely accuracy, MCC, Kappa, G-mean, ROC-AUC, CBA, IAM, P4, WCA, WRA, ACD, H, H informed, WA, and EWA 
exhibit  symmetry under relabeling of class $+$ and $-$, reflected in the heatmaps as qualitative symmetry under the transformation $(r_+, r_C) \to (1-r_+, 1-r_C)$%
\footnote{The symmetry is qualitative, as the asymmetric cost distributions break exact symmetry.}.
While symmetry is often considered desirable, our focus here is on robustness of correlation with TCC. Thus, we classify metrics based on the qualitative pattern of their correlation.

\subsubsection{Anti-Diagonal Robust Metrics}

As discussed in Section \ref{sec:cost_based_reweighting} for example-independent costs, when the ratio of positives in the development and target dataset is the same, standard accuracy is perfectly consistent with TCC if $r_C=1/2$.
However, we can also expect a good alignment with TCC when the majority class has the larger misclassification cost (i.e., $r_C$ and $r_+$ are both small or both large): in this case balancing techniques are counterproductive.
Indeed, in these regions, accuracy outperforms many sophisticated metrics, particularly those designed to handle class imbalance \emph{per se}, without considering the UCC ratio.
Besides standard accuracy, also CBA, IAM, WCA, and H exhibit strong correlation with TCC close to the anti-diagonal of the heatmap (i.e., the line $r_C=r_+$).
More precisely, accuracy, CBA, IAM, and H diverge significantly from TCC in two regions: (i) $r_+ \lesssim 0.5$ with $r_C \gtrsim 1-r_+$ and (ii) $r_+ \gtrsim 0.5$ with $r_C \lesssim 1-r_+$.
WCA exhibits a slightly different pattern,
diverging from TCC in 
(i) $r_+ \lesssim 0.5$, with $0.5 \lesssim r_C \lesssim 1-r_+$ and 
(ii) $r_+ \gtrsim 0.5$, with $1-r_+ \lesssim r_C \lesssim 0.5$.
Standard and weighted Spearman coefficients (Figures \ref{fig:heatmap_churn_standard}, \ref{fig:heatmap_credit_standard}, \ref{fig:heatmap_churn_weighted}, and \ref{fig:heatmap_credit_weighted}) display analogous patterns for this group of metrics.

\subsubsection{Main Diagonal Symmetric Metrics}
\label{sec:main_diagonal_metrics}

Several metrics exhibit qualitatively symmetric or antisymmetric correlation patterns with respect to reflection along the main diagonal ($r_C=1-r_+$).
According to the findings in Section \ref{sec:handling_class_imbalance}, metrics designed to compensate for class imbalance without accounting for UCCs are expected to show good correlation with TCC only for $r_C\simeq 1-r_+$, i.e., close to the main diagonal. This pattern is observed for informedness, markedness, MCC, Cohen’s Kappa, WRA and ROC-AUC (single parameter), which we thus refer to as \emph{rebalancing metrics}. In the weighted case, the decline in correlation away from the diagonal is more pronounced for informedness, ROC-AUC, WRA, and even stronger for Kappa.

Recall, F1, and B-ROC show instead strong correlation with TCC above the main diagonal and strong anticorrelation below (less pronounced for F1 in the standard correlation), while specificity shows the opposite behavior.
Precision and NPV behave differently depending on whether the standard or weighted correlation coefficient is used: in the former case (Figures \ref{fig:heatmap_churn_standard} and \ref{fig:heatmap_credit_standard}), they behave as rebalancing metrics, with good correlation along the diagonal, which fades toward uncorrelation or even anticorrelation at the extremes of $r_+$. When considering the weighted correlation coefficient (Figures \ref{fig:heatmap_churn_weighted} and \ref{fig:heatmap_credit_weighted}), the line of maximal correlation shifts slightly above (for precision) or below (for NPV), with the diagonal itself dividing the heatmap into two regions qualitatively similar to recall and specificity.
P4 and G-mean, when analyzed with the standard correlation (Figures \ref{fig:heatmap_churn_standard} and \ref{fig:heatmap_credit_standard}), behave as the rebalancing metrics, with overall lower correlation. Under the weighted correlation (Figures \ref{fig:heatmap_churn_weighted} and \ref{fig:heatmap_credit_weighted}), they exhibit a reversed pattern, with low correlation along the diagonal and stronger correlation elsewhere.

\subsubsection{Globally Robust Metrics}

ACD, H informed, WA, and EWA are the top-performing metrics, showing strong correlation with TCC across nearly all scenarios. ACD performs slightly worse than the others in this group, particularly under the weighted correlation (Figures \ref{fig:heatmap_churn_weighted} and \ref{fig:heatmap_credit_weighted}), where it exhibits weaknesses qualitatively similar to those of accuracy, CBA, IAM, and H.
The robustness of this group of metrics was to some extent expected, as
they incorporate the example-independent TCC (Eq.~\ref{eq:total_cost}, with ${\mathrm{TCC}}_{\mathrm{min}} = 0$). However, Figures \ref{fig:heatmap_churn_standard}, \ref{fig:heatmap_credit_standard}, \ref{fig:heatmap_churn_weighted}, and \ref{fig:heatmap_credit_weighted} show robust and consistent correlation with the example-dependent cost (Eq.~\ref{eq:total_cost_example_dep_3}), indicating that the fluctuation terms are negligible in this scenario.
We omit the heatmaps for MSU and C-score because they are linearly related to WA.

\subsection{Validity Limits of the Example-Independent Approximation}
\label{sec:extreme_statistics}

To quantitatively assess the limits of the example-independent approximation (see also Section~\ref{sec:example_dependent_TCC}), we consider highly skewed, long-tailed distribution of UCCs.
This scenario builds from the first use case (churn prediction) described in Section~\ref{sec:use_cases}, by introducing a subset of \emph{massive customers}, whose revenue dominates the overall distribution.
Specifically, we define $N_{\mathrm{mc}}$ massive customers whose combined revenue accounts for a fraction $f_{\mathrm{r}}$ of the total revenue.

Two quantities characterize the degree of skewness and tail heaviness of the revenue distribution: the fraction of massive customers
$$
f_{\mathrm{mc}} = \frac{N_{\mathrm{mc}}}{N_{\mathrm{tot}}}
$$
  and the fraction $f_{\mathrm{r}}$ of the total revenue generated by these customers.
We therefore analyze a grid of $f_{\mathrm{r}}$, $f_{\mathrm{mc}}$ values, with $f_{\mathrm{r}}$ ranging from $20\%$ to $99\%$ and $f_{\mathrm{mc}}$ ranging from $1\%$ to $20\%$.

%
\begin{figure}[ht]
    \centering
    \includegraphics[width=\columnwidth]{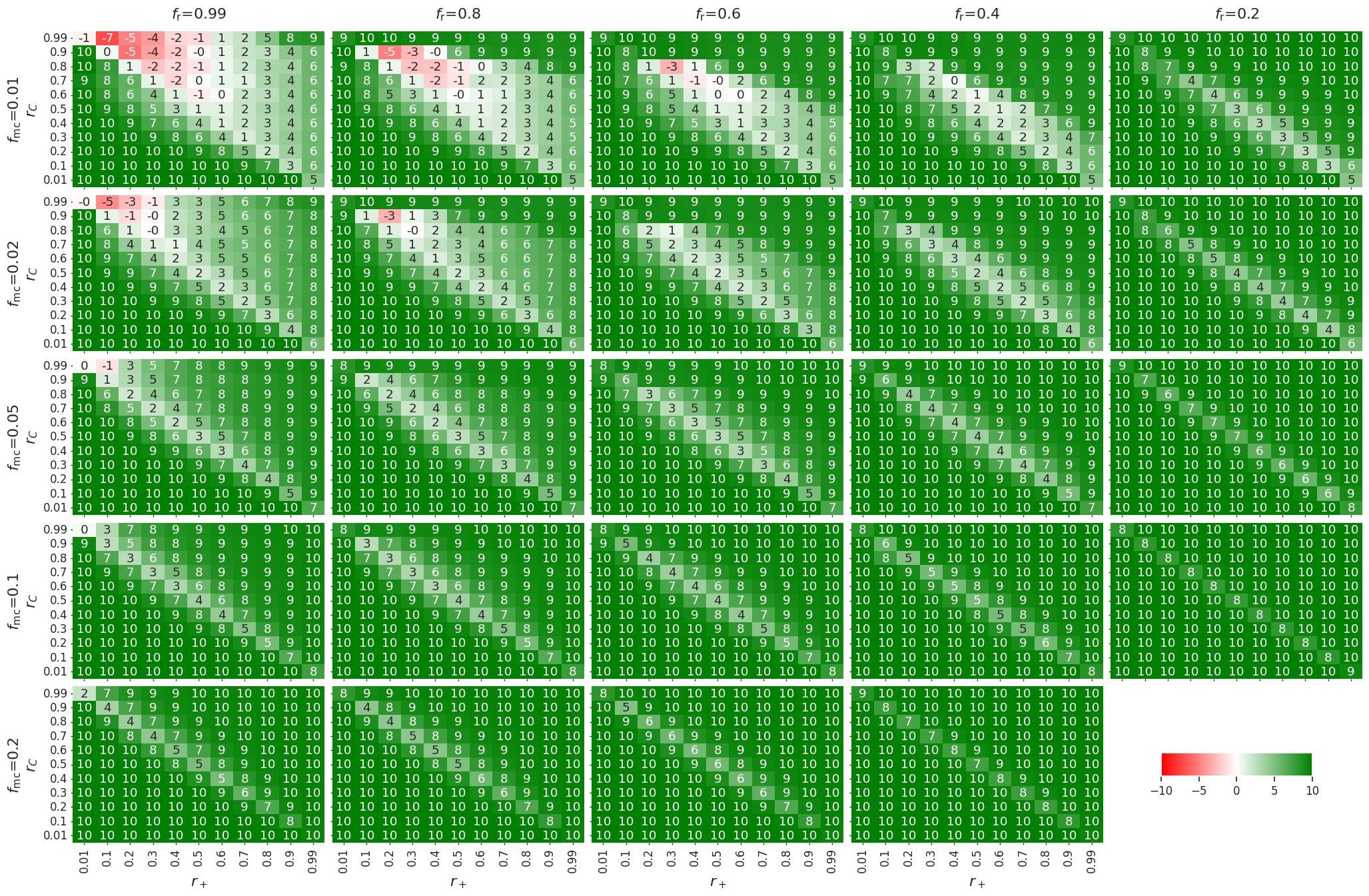}
    \caption{Heatmaps showing {\bf standard Spearman} correlation between TCC and WA over a grid of values for $r_+$ (horizontal axis) and $r_C$ (vertical axis) for the {\bf churn prediction with extreme statistics} use case. Results are reported for different values of the fraction $f_{\mathrm{mc}}$ of massive customers and the fraction $f_{\mathrm{r}}$ of total revenue associated with them. 
    To improve readability, correlation coefficients are multiplied by 10 and rounded.}
    \label{fig:heatmap_churn_extreme_standard}
\end{figure}
%

Figure \ref{fig:heatmap_churn_extreme_standard} reports the standard Spearman correlation between TCC and WA for different combinations of $f_{\mathrm{r}}$ and $f_{\mathrm{mc}}$.
As expected, WA exhibits increasing discrepancies with TCC as the revenue distribution becomes more concentrated and heavy-tailed; i.e., these discrepancies are significant for large values of $f_{\mathrm{r}}$ ($\gtrsim 0.6$) and small values of $f_{\mathrm{mc}}$ ($\lesssim 0.02$).

Conversely, when the revenue fraction associated with the massive customers is more moderate ($f_{\mathrm{r}}\lesssim 0.4$) or the fraction of massive customers is sufficiently large ($f_{\mathrm{mc}}\gtrsim 0.05$), WA maintains relatively strong correlation and near-consistent behavior with respect to TCC. These experiments provide therefore an empirical estimation of the validity limits of WA, and more generally of confusion-matrix-based evaluation metrics, in the presence of example-dependent costs with highly skewed distributions.
The procedure to explore the $r_C$, $r_+$ parameter grid is the same as that described in Section \ref{sec:data_algorithm} and the code is contained in the same repository.

\subsection{Practical Guidelines for Classifier Evaluation}

The proposed evaluation framework is applicable across all stages in which classifier performance is assessed: primarily model validation and testing, but also model training.

\subsubsection{Model Training}
\label{sec:framework_for_model_training}

Model training typically relies on surrogate loss functions that are differentiable and, in many cases,  convex. Common choices for binary classification include 
binary cross entropy, 
hinge loss,
squared loss, and
exponential loss.
All these loss functions can be expressed in the form of Eq.~\ref{eq:kpi_average} with uniform weights $v_a=1/N_{\mathrm{tot}}$; their per-example contributions are 
\begin{itemize}
  \item binary cross entropy: $k_a=- [y_a \log(p_a) + (1- y_a)\log(1-p_a)]$;
  \item hinge loss: $k_a = \mathrm{max}(0, 1-y_ap_a)$;
  \item squared loss: $k_a = (y_a-p_a)^2$;
  \item exponential loss: $k_a=\mathrm{exp}(-y_ap_a)$. 
\end{itemize}
Here, $y_a$ denotes the true label of example $a$ (0 or 1) and $p_a$ the corresponding model output, e.g., the predicted probability  of $a$ being in positive class.

The reweighting framework introduced in Sections \ref{sec:handling_class_imbalance}, \ref{sec:cost_based_reweighting}, and \ref{sec:target_set_balancing} can therefore be directly applied to these loss functions, enabling principled handling of class imbalance and unequal UCCs.

This requires estimating the ratio of positives $r_+^{\mathrm{t}}$ in the target dataset and the UCCs ratio $r_C$. 
The latter can be estimated using the procedure described in Appendix \ref{app:weight_estimation};
conversely, when the target distribution is unavailable, $r_+^{\mathrm{t}}$ can be approximated by the positive fraction $r_+$ observed in the training set, provided that no substantial distribution shift is expected between the development and target datasets.
The reweighted loss function is obtained by rescaling the original weights $v_a$ according to Eq.~\ref{eq:w_target_kpi_unequal_UCCs}.

If the original loss is consistent with TCC minimization under balanced conditions and equal UCCs, the previously discussed reweighted loss remains consistent in the more general setting characterized by class imbalance ($r_+^{\mathrm{t}}\neq r_+$) and asymmetric misclassification costs ($r_C\neq 1/2$).

\subsubsection{Model Validation and Testing}
\label{sec:TCC_after_validation}

During model validation and testing, the results presented in Section \ref{sec:results} indicate that WA (or EWA in the presence of substantial uncertainty in UCCs) is well aligned with TCC minimization. This holds not only for example-independent UCCs, for which the equivalence is formally established in Eq.~\ref{eq:WA_propto_TCC}, but also for example-dependent UCCs, provided that their distribution does not exhibit extremely long tails (see Section \ref{sec:extreme_statistics}).

In contrast, the use of alternative metrics, included several widely adopted evaluation measures, may lead to sub-optimal model selection, with a potentially significant impact in terms of TCC;
in the following, we quantify this effect.

Consider a validation setting in which classifier hyperparameters must be selected. 
To emulate model performance across different hyperparameter configurations, we assume a parametric relationship between the True Positive Rate (TPR, i.e., $\mathrm{TP}/\mathrm{P}$) and the False Positive Rate (FPR, i.e., $\mathrm{FP}/\mathrm{N}$), namely  $\mathrm{TPR} = (\mathrm{FPR})^2$, which corresponds to an area under the ROC curve of $2/3$.
The candidate models are generated from this FPR-TPR relationship, by uniformly sampling FPR in the interval [0,1] with step size $0.01$.

We analyze four representative scenarios:
\begin{enumerate}
  \item mild class imbalance ($r_+=0.2$) with substantially higher cost associated with false positives ($r_C=0.01$);
  \item mild imbalance ($r_+=0.2$) with substantially higher cost associated with false negatives  ($r_C=0.99$);
  \item strong imbalance ($r_+=0.01$) with higher cost associated with false negatives ($r_C=0.9$);
  \item strong imbalance ($r_+=0.01$) with higher cost associated with false positives ($r_C=0.1$).
\end{enumerate}

For each scenario, we select, from the previously described set of candidates, the optimal model according to each evaluation metric $X$ under consideration.
We then compute the performance gap $\Delta \mathrm{TCC}$, defined as the difference between the TCC of the model selected by metric $X$ and the minimum achievable TCC among all candidate models.
Therefore, $\Delta \mathrm{TCC}$ quantifies the economical cost induced by the sub-optimal classifier performance, due to the use of metric $X$ during model selection.

To incorporate example-dependent UCCs, we adopt the cost structure introduced in the first use case in Section \ref{sec:use_cases}; for each configuration, the sets of true positives and false positives are generated via uniform random sampling consistent with the specified TPR and FPR values. The computation of $\Delta \mathrm{TCC}$ is repeated over 1000 independent trials, and Table \ref{tab:TCC_after_validation} reports the corresponding mean values.

\begin{table}[ht]
\centering
 \caption{For each metric $X$, we provide the difference between the TCC of the model selected by optimizing $X$ and the minimal TCC in the four validation scenarios described in Section \ref{sec:TCC_after_validation}.}
\label{tab:TCC_after_validation}
\vspace{2mm}
\begin{tabular}{c|cccc}
\hline
Metric       & Scenario 1 & Scenario 2 & Scenario 3 & Scenario 4 \\ \hline
Accuracy     & 0          & 591        & 7          & 0          \\
CBA          & 795        & 565        & 7          & 0          \\
IAM          & 795        & 565        & 7          & 0          \\
H            & 0          & 591        & 7          & 0          \\
WCA          & 0          & 1          & 253        & 0          \\
Kappa        & 0          & 591        & 253        & 3155       \\
Informedness & 0          & 591        & 253        & 3155       \\
ROC-AUC      & 0          & 591        & 253        & 3155       \\
WRA          & 0          & 591        & 253        & 3155       \\
MCC          & 0          & 591        & 253        & 3155       \\
Markedness   & 3842       & 17         & 253        & 3155       \\
Precision    & 3916       & 1          & 253        & 3155       \\
NPV          & 0          & 591        & 253        & 3155       \\
P4           & 2778       & 292        & 253        & 3155       \\
G-mean       & 2778       & 292        & 154        & 1853       \\
Jaccard      & 3916       & 1          & 253        & 3155       \\
F1           & 3916       & 1          & 253        & 3155       \\
Recall       & 3916       & 1          & 253        & 3155       \\
B-ROC        & 3916       & 1          & 253        & 3155       \\
Specificity  & 0          & 591        & 7          & 0          \\ 
ACD          & 0          & 110        & 7          & 0          \\
WA           & 0          & 1          & 7          & 0          \\
H informed   & 0          & 1          & 7          & 0          \\
EWA          & 0          & 1          & 7          & 0  \\
\hline       
\end{tabular}
\end{table}

Most metrics exhibit large $\Delta \mathrm{TCC}$ values in at least one scenario, indicating that their use for model selection may lead to substantial economical losses. In contrast, only a limited subset of metrics consistently yields low $\Delta \mathrm{TCC}$ across all analyzed scenarios, namely
 the globally robust metrics, in particular WA, H informed, and EWA; ACD, while generally competitive, exhibits a non-negligible $\Delta \mathrm{TCC}$ in Scenario 2. 
 
 Since minimizing $\Delta \mathrm{TCC}$ is equivalent to maximizing RoI, these results also indicate that using WA, H informed, or EWA for model selection yields the highest RoI across the analyzed scenarios.

\section{Conclusion}

In Section \ref{sec:WA_and_framework} we introduced WA, an  evaluation metric for binary classifiers that can be interpreted as a cost-consistent weighted version of accuracy, together with a reweighting framework -- applicable to any metric or loss function that, like WA, can be expressed as a linear combination of example-dependent quantities -- 
for handling class imbalance in cost-sensitive settings without relying on resampling techniques. Within this framework, we established several key results for the case of example-independent UCCs:
\begin{itemize}
 \item in Section \ref{sec:WA} we proved that maximizing WA is equivalent to minimizing TCC;
 \item in Section \ref{sec:cost_based_reweighting}, we derived the condition under which metrics and rebalancing techniques designed to compensate class imbalance remain coherent with TCC minimization, namely $r_C\simeq 1-r_+$ (Eq.~\ref{eq:rescaling_regime});
 \item in Section \ref{sec:target_set_balancing}, we generalized the framework to handle \emph{development} and \emph{target} datasets with different ratios of positives and discussed the corresponding validity limits of standard accuracy.
\end{itemize}

Since most real-world applications involve example-dependent UCCs, Section \ref{sec:metrics_comparison_with_TCC} focused on realistic use cases  characterized by heterogeneous costs,  systematically exploring the full range of possible values for the cost ratio ($r_C$) and the ratio of positives ($r_+$), and analyzing the correlation between several evaluation metrics and TCC.

The empirical behavior of rebalancing metrics observed in Section \ref{sec:main_diagonal_metrics} qualitatively extends to a broad range of example-dependent scenarios the theoretical condition derived for example-independent UCCs, namely that rebalancing-based approaches remain coherent with TCC minimization only near the regime $r_C\simeq 1-r_+$ (Eq.~\ref{eq:rescaling_regime}). More generally, this result suggests that widely adopted methods for handling class imbalance, including standard undersampling and oversampling techniques, implicitly rely on assumptions about the relationship between class imbalance and misclassification costs that may not hold in realistic applications.

Among the analyzed confusion-matrix-based metrics, only H informed, EWA, WA, C-score, and MSU exhibit robust correlation with TCC across all tested scenarios, thereby reducing the risk of misleading model assessments.
H informed and EWA require the additional complexity of a probabilistic formulation, which increases computational cost without providing clear advantages in the analyzed scenarios. 
WA, in contrast, combines robust empirical behavior with a straightforward interpretation, cross-dataset comparability, and a natural extension to scenarios in which the target and development datasets differ; these properties make WA suitable both for comparing models evaluated on different datasets and for validation-based hyperparameter tuning.
In Section \ref{sec:TCC_after_validation}, we further quantified the practical impact of metric choice by measuring the RoI advantage obtained  when WA is used for model validation instead of alternative evaluation metrics.

The experimental results in Sections \ref{sec:results} also indicate that the example-independent approximation of TCC, obtained by neglecting the fluctuation terms in Eq.~\ref{eq:total_cost_example_dep_3}, remains consistent and reliable in all examined scenarios.
This observation is particularly relevant because such an approximation is implicitly assumed by any evaluation framework based on the confusion matrix formalism.
Motivated by this consideration, Section \ref{sec:extreme_statistics} systematically investigated the validity limits of this approximation in the presence of highly skewed and heavy-tailed cost distributions, i.e., scenarios in which a very small fraction of \emph{massive examples} accounts for a relevant fraction of the total cost. The results show that near-consistent behavior is still observed when either the fraction of total cost associated with the massive examples remains moderate or the fraction of examples being \emph{massive} is sufficiently large.

\begin{appendices}

\section{Estimation of the WA Weight}\label{app:weight_estimation}

In an ideal scenario, the UCCs defined in  Section \ref{sec:related_work_cost} can be assessed by estimating the cost associated to the actions taken in response to each type of classification outcome (false positive, true positive, etc.). Note that the cost matrix may include opportunity costs, i.e., foregone benefits due to missed opportunities, as in the examples described in Section \ref{sec:use_cases}; costs or benefits can be measured relatively to any baseline. However, the baseline must remain fixed, i.e., the reference point for zero cost should not change~\cite{cost_sensitive2001foundations}.

In practice, estimating the full confusion matrix may be challenging. Therefore, we often focus on the more attainable goal of estimating the weight $w$ in Eq.~\ref{eq:w_accuracy_binary}, or equivalently, the UCC ratio $\rho=C_{\mathrm{FN}} / C_{\mathrm{FP}}$, which suffices for computing WA in Eq.~\ref{eq:w_accuracy_binary}.
We propose therefore a procedure to estimate the weight range $w_{\mathrm{min}} \leq w \leq w_{\mathrm{max}}$.
Here, we assume that the development and target datasets have the same ratio of positive examples; the generalization to $r_+ \neq r_+^{\mathrm{t}}$ is described in Eq.~\ref{eq:w_target_accuracy}.

\subsection{Unit Misclassification Costs Ratio}

The first approach to estimating the weight range is via the UCC ratio $\rho$ -- i.e., estimating how many false positives are, on average, equivalent to a single false negative -- and using the equivalence $w = \rho/(\rho+1)$. 
For instance, in bankruptcy prediction, a false negative -- where an auditor incorrectly assesses a company as solvent -- may result in liability to creditors and shareholders. Conversely, a false positive -- where a solvent company is incorrectly assessed as insolvent -- may lead to reduced access to credit and increased uncertainty. Based on these considerations, \cite{cost_sensitive1977zetatm} and \cite{cost_sensitive2000comparison} estimated $\rho \simeq 35$ (i.e., $w \simeq 0.97$) and $10 \lesssim \rho \lesssim 50$ (i.e., $0.91 \lesssim w \lesssim 0.98$), respectively.

\subsection{Constraints from Ranking of Emblematic Models}

If the previous approach is not feasible, an alternative method is to infer constraints on $w$ by ranking a set of emblematic classification results. Even if we cannot estimate the costs $C_{\mathrm{FP}}$ and $C_{\mathrm{FN}}$ (or their ratio), we may still be able to determine whether one specific outcome is more costly than another.

\subsubsection{Construction of the Emblematic Model Set}

The choice of emblematic models may depend on the use case and the ratio of positive examples. Ideally, we should select a small set of models, sufficiently simple to avoid overcomplexity in the ranking of the outcomes, yet diverse enough to yield meaningful constraints on $w$. As a baseline, we consider the models described in Table \ref{tab:relevant_CM_accuracy}:  
(i) $\mathcal{M}_+$ and $\mathcal{M}_-$, which always predict the same class;  
(ii) $\mathcal{M}_{\mathrm{bad}}$, which misclassifies a fraction $\alpha$ of examples in both classes;  
(iii) $\mathcal{M}_{\mathrm{bad-}}$ and $\mathcal{M}_{\mathrm{bad+}}$, which misclassify a fraction $\alpha$ of negatives or positives, respectively, while perfectly classifying the other class.

%
%
\begin{table}[ht]
\centering
\footnotesize
 \caption{Set of emblematic models to determine constraints on $w$. $A_{\mathrm{num}}$ is the numerator in Eq.~\ref{eq:w_accuracy_binary}.}
\label{tab:relevant_CM_accuracy}
\begin{tabular}{ccc}
\toprule
 Name &Description & $A_{\mathrm{num}}$\\
\midrule
 $\mathcal{M}_+$ & Dummy model always predicting class $+$. & $w\mathrm{P}$\\
 $\mathcal{M}_-$ & Dummy model always predicting class $-$. & $(1-w) \mathrm{N}$\\ 
 $\mathcal{M}_{\mathrm{bad}}$ & Misclassifies a fraction $\alpha$ of examples in both classes. & $(1-\alpha)[w\mathrm{P} +(1-w)\mathrm{N}]$ \\
 $\mathcal{M}_{\mathrm{bad-}}$ & Misclassifies a fraction $\alpha$ of negatives, perfect on positives. & $w\mathrm{P} + (1-\alpha)(1-w)\mathrm{N}$ \\
 $\mathcal{M}_{\mathrm{bad+}}$ & Misclassifies a fraction $\alpha$ of positives, perfect on negatives. & $(1-\alpha)w\mathrm{P} + (1-w)\mathrm{N}$\\
\bottomrule
\end{tabular}
\end{table}
%
%

The parameter $\alpha$, representing the fraction of misclassified examples, can be chosen \emph{ad hoc}
to facilitate model ranking. To derive meaningful constraints on $w$, we recommend choosing $\alpha$ in the range $0.5 \lesssim \alpha \lesssim 0.75$.

\subsubsection{Ranking of Emblematic Models}

To rank the models previously identified, we must understand the use case from a business perspective. For example, is it preferable to correctly identify all positives (as in $\mathcal{M}_{\mathrm{bad-}}$) or all negatives (as in $\mathcal{M}_{\mathrm{bad+}}$)?
For illustrative purposes, let us assume a ratio $r_+=0.05$ of positive examples, and let us use the models from Table \ref{tab:relevant_CM_accuracy} with $\alpha = 0.6$. Suppose it is impractical to react to a large number of predicted positives (e.g., contacting many customers in churn prediction, treating many patients in cancer detection, etc.). In this scenario, the worst outcome is likely $\mathcal{M}_+$, and we expect $A(\mathcal{M}_{\mathrm{bad-}}) \lesssim A(\mathcal{M}_{\mathrm{bad+}})$.

Moreover, $\mathcal{M}_-$, which predicts only negatives, may be slightly preferable to $\mathcal{M}_{\mathrm{bad}}$, since its output could still be useful if a future positive prediction occurs. To complete the ranking, we assume $\mathcal{M}_{\mathrm{bad-}}$ is preferable to $\mathcal{M}_-$, as the former misclassifies a fraction $\alpha$ of negatives and no positive example, and therefore produces a more informative output compared to the latter, which misclassifies all positives.

In summary, we obtain the following ranking:
\begin{equation}
    A(\mathcal{M}_+) \lesssim 
    A(\mathcal{M}_{\mathrm{bad}}) \lesssim 
    A(\mathcal{M}_-) \lesssim
    A(\mathcal{M}_{\mathrm{bad-}}) \lesssim
    A(\mathcal{M}_{\mathrm{bad+}}).
\end{equation}

\subsubsection{Inferring the Constraints from the Ranking}

Since $\mathrm{P}$ and $\mathrm{N}$ are fixed for a given dataset, ranking the models by WA is equivalent to ranking them by the numerator $A_{\mathrm{num}}$ in Table \ref{tab:relevant_CM_accuracy}.
For $\alpha\geq 0.5$, the 
resulting constraints on $w$ are:
\begin{equation}\label{eq:w_constraints_from_ranking}
    \left[1 +\frac{\mathrm{P}}{\alpha\, \mathrm{N}}\right]^{-1} \lesssim w \lesssim \left[1+ \frac{\alpha\, \mathrm{P}}{(1-\alpha)\mathrm{N}}\right]^{-1}.
\end{equation}
For $\alpha =0.6$ and $r_+=0.05$, Eq.~\ref{eq:w_constraints_from_ranking} becomes $0.919 \lesssim w \lesssim 0.927$,
where the upper bound follows from the condition $A(\mathcal{M}_+) \lesssim A(\mathcal{M}_{\mathrm{bad}})$, and the lower bound from $A(\mathcal{M}_{\mathrm{bad-}}) \lesssim A(\mathcal{M}_{\mathrm{bad+}})$.

\section{Use Cases Details}\label{app:use_cases_details}

\subsection{Churn Prediction}\label{app:churn}
  
Each customer $a$ predicted as positive causes a cost $M$ for the retention measure, which includes the time and effort of the commercial team to contact the customer and implement the measure. Here we assume that this retention cost is example-independent, so $e_a^{\mathrm{FP}} = d_a^{\mathrm{TP}} = M$.  
Since the experiment described in Section \ref{sec:data_algorithm} requires spanning all possible UCC ratios $r_C$, we tune the cost of the retention measure $M$ accordingly.  

Conversely, for a customer $a$ predicted as negative, the cost (or missing income) is $R_a P_{\mathrm{eff}}$, where $P_{\mathrm{eff}}$ is the probability of effectiveness of the retention measure (assumed example-independent and fixed to $25\%$), while $R_a$ is the revenue the company would have earned from $a$ had the customer not churned, and is assumed proportional to the customer's revenue.

To avoid negative values on the false negative costs, we clip each $D_a^{\mathrm{FN}}$ to positive values, i.e., we set $D_a^{\mathrm{FN}} = {\rm max}(0, R_a\, P_{\mathrm{eff}}-M)$. 
To obtain the desired value for $r_C$, we must solve the equation $M = \gamma\sum_a{\rm max}(M,P_{\mathrm{eff}}R_a)$, with $\gamma=(1-r_C)/N_{\mathrm{tot}}$, whose solution is $M=\gamma\sum_{a=A+1}^{N_{\mathrm{tot}}}P_{\mathrm{eff}}R_a / (1-\gamma A)$, subject to the constraint $P_{\mathrm{eff}}R_A \leq M \leq P_{\mathrm{eff}}R_{A+1}$.
Under these assumptions, the average UCCs are $C_{\mathrm{FP}} = M$ and $C_{\mathrm{FN}} = R_{\mathrm{avg}}^{\mathrm{clip}} P_{\mathrm{eff}} - M$, where $R_{\mathrm{avg}}^{\mathrm{clip}}$ denotes the expected value of $R_a^{\mathrm{clip}}={\rm max}(R_a,M/P_{\mathrm{eff}})$ over the target dataset, while the fluctuations are $\delta_a = P_{\mathrm{eff}}(R_a^{\mathrm{clip}} - R_{\mathrm{avg}}^{\mathrm{clip}})$ and $\epsilon_a=0$.

\subsection{Credit Scoring}\label{app:credit}

For credit scoring, we use the approach described in \cite{credit2014example}.
In that framework, the costs of correct classifications, $d_a
^{\mathrm{TP}}$ and $e_a^{\mathrm{TN}}$, are assumed to be zero for every customer $a$. 
The loss $D_a^{\mathrm{FN}}=d_a^{\mathrm{FN}}$ if the customer $a$ defaults is proportional to their credit line, while the cost $E_a^{\mathrm{FP}}=e_a^{\mathrm{FP}}$ of a false positive is the sum of two financial components, $R_a$ and $G_{\mathrm{avg}}$.
The first term is the loss of profit from rejecting a customer who would have repaid the loan; it depends on the loan parameters (see \cite{credit2014example} for details). 
Differently from \cite{credit2014example}, we introduce clipping of the customer debt ratio to the interval $[0,1]$, ensuring data consistency.

The term $G_{\mathrm{avg}}$ reflects the assumption that the financial institution
does not keep the capital of a rejected customer idle, but instead allocates it to an alternative customer. 
Since no additional information about this alternative customer is available, 
we assume that the customer defaults with probability equal to the prior positive rate $r_+$. 
Under this assumption, 
$G_{\mathrm{avg}}=-R_{\mathrm{avg}}\cdot (1-r_+)+\mathrm{Cl}\cdot L_{\mathrm{gd}}\cdot r_+$, 
where $\mathrm{Cl}$ is the average credit line, $R_{\mathrm{avg}}$ the average profit, and $L_{\mathrm{gd}}$ the loss-given-default ratio, i.e., the fraction of the customer’s debt that is lost in case of default.

The parameter used to tune the UCC ratio $r_C$ is $L_{\mathrm{gd}}$. To ensure that $L_{\mathrm{gd}}$ remains non-negative, 
the analysis must be restricted to the domain $r_C < 1 / (1 + r_+)$, hence the missing values in the top right corner of each heatmap in Figures \ref{fig:heatmap_credit_standard} and \ref{fig:heatmap_credit_weighted}.

\end{appendices}

\bibliographystyle{plain}

\bibliography{sn-bibliography}

@article{imbalance2002smote,
  title={SMOTE: synthetic minority over-sampling technique},
  author={Chawla, Nitesh V and Bowyer, Kevin W and Hall, Lawrence O and Kegelmeyer, W Philip},
  journal={Journal of artificial intelligence research},
  volume={16},
  pages={321--357},
  year={2002}
}

@inproceedings{imbalance2003c4,
  title={C4.5, class imbalance, and cost sensitivity: why under-sampling beats over-sampling},
  author={Drummond, Chris and Holte, Robert C},
  booktitle={Int. Conf. on Mach. Learn.},
  volume={11},
  year={2003}
}

@Inbook{imbalance2010data,
  title={Data mining for imbalanced datasets: An overview},
  author={Chawla, Nitesh V},
  booktitle={Data Min. and Know. Disc. handbook},
  pages={853--867},
  year={2005},
  publisher={{Springer US}},
  address={Boston, MA},
  editor={Maimon, Oded and Rokach, Lior},
  doi={10.1007/0-387-25465-X_40}
}

@article{imbalance2017imbalanced,
  title={Imbalanced-learn: A python toolbox to tackle the curse of imbalanced datasets in Mach. Learn.},
  author={Lema\^{i}tre, Guillaume and Nogueira, Fernando and Aridas, Christos K},
  journal={Jour. of Mach. Learn. Res.},
  volume={18},
  number={17},
  pages={1--5},
  year={2017}
}

@article{imbalance2018curse,
  title={The curse of class imbalance and conflicting metrics with machine learning for side-channel evaluations},
  author={Picek, Stjepan and Heuser, Annelie and Jovic, Alan and others},
  journal={IACR Trans. on Cryptographic Hardware and Embedded Syst.},
  volume={2019},
  number={1},
  pages={209--237},
  year={2018},
  doi={10.13154/tches.v2019.i1.209-237}
}

@inproceedings{imbalance2020streams,
  title={Cost-sensitive learning for imbalanced data streams},
  author={Loezer, Lucas and Enembreck, Fabr{\'\i}cio and Barddal, Jean Paul and others},
  booktitle={Proc. of the 35th annual ACM symposium on applied Comp. },
  pages={498--504},
  year={2020},
  doi={10.1145/3341105.3373949}
}

@article{imbalance2021balancing,
  title={The balancing trick: Optimized sampling of imbalanced datasets—A brief survey of the recent State of the Art},
  author={Susan, Seba and Kumar, Amitesh},
  journal={Eng. Reports},
  volume={3},
  number={4},
  pages={e12298},
  year={2021},
  publisher={Wiley Online Library},
  doi={10.1002/eng2.12298}
}

@article{imbalance2021performance,
  title={Performance analysis of cost-sensitive learning methods with application to imbalanced medical data},
  author={Mienye, Ibomoiye Domor and Sun, Yanxia},
  journal={Artif Intell Rev},
  volume={57},
  number={80},
  year={2024},
  publisher={Springer},
  doi={10.1007/s10462-023-10652-8}
}

@inproceedings{metrics1999rule,
  title={Rule evaluation measures: A unifying view},
  author={Lavra{\v{c}}, Nada and Flach, Peter and Zupan, Blaz},
  booktitle={Int. Conf. on inductive logic programming},
  pages={174--185},
  year={1999},
  organization={Springer},
  doi={10.1007/3-540-48751-4_17}
}

@article{metrics2014strategy,
  title={A strategy on selecting performance metrics for classifier evaluation},
  author={Liu, Yangguang and Zhou, Yangming and Wen, Shiting and others},
  journal={Int. Jour. of Mobile Comp.  and Multimedia Communications (IJMCMC)},
  volume={6},
  number={4},
  pages={20--35},
  year={2014},
  publisher={IGI Global},
  doi={10.4018/IJMCMC.2014100102}
}

@article{metrics2015review,
  title={A review on evaluation metrics for data classification evaluations},
  author={Hossin, Mohammad and Sulaiman, Md Nasir},
  journal={Int. Jour. of Data Min. \& Know. Manag. process},
  volume={5},
  number={2},
  pages={1},
  year={2015},
  publisher={Academy \& Industry Res. Collaboration Center (AIRCC)},
  doi={10.5121/ijdkp.2015.5201}
}

@article{metrics2011evaluation,
  title={Evaluation: From Precision, Recall and {F}-Measure to {ROC}, Informedness, Markedness \& Correlation},
  author={Powers, David},
  journal={Jour. of Mach. Learn. Tech.},
  volume={2},
  number={1},
  pages={37--63},
  year={2011},
  publisher={BioInfo Publications}
}

@inproceedings{metrics2023metrics,
  title={Metrics for evaluating classification algorithms},
  author={Muntean, Mihaela and Militaru, Florin-Daniel},
  booktitle={Education, Res. and Business Tech.: Proc. of 21st Int. Conf. on Informatics in Economy (IE 2022)},
  pages={307--317},
  year={2023},
  organization={Springer},
  doi={10.1007/978-981-19-6755-9_24  }
}

@article{metrics2024evaluation,
  title={Evaluation metrics and statistical tests for machine learning},
  author={Rainio, Oona and Teuho, Jarmo and Kl{\'e}n, Riku},
  journal={Scientific Reports},
  volume={14},
  number={1},
  pages={6086},
  year={2024},
  publisher={Nature Publishing Group UK London},
  doi={https://doi.org/10.1038/s41598-024-56706-x}
}

@inproceedings{new_metrics2006b,
  title={{B-ROC} curves for the assessment of classifiers over imbalanced data sets},
  author={C{\'a}rdenas, Alvaro A and Baras, John S},
  booktitle={Proc. of the national Conf. on Art. Int. },
  volume={21},
  number={2},
  pages={1581},
  year={2006}
}

@article{new_metrics2006mean,
  title={The mean subjective utility score, a novel metric for cost-sensitive classifier evaluation},
  author={McDonald, Ross A},
  journal={Pattern Recognition Lett.},
  volume={27},
  number={13},
  pages={1472--1477},
  year={2006},
  publisher={Elsevier},
  doi={10.1016/j.patrec.2006.02.012}
}

@article{new_metrics2009measuring,
  title={Measuring classifier performance: a coherent alternative to the area under the {ROC} curve},
  author={Hand, David J},
  journal={Mach. Learn.},
  volume={77},
  number={1},
  pages={103--123},
  year={2009},
  publisher={Springer},
  doi={10.1007/s10994-009-5119-5}
}

@article{new_metrics2012novel,
  title={A novel profit maximizing metric for measuring classification performance of customer churn prediction models},
  author={Verbraken, Thomas and Verbeke, Wouter and Baesens, Bart},
  journal={IEEE Trans. on Know. and Data Eng.},
  volume={25},
  number={5},
  pages={961--973},
  year={2013},
  doi={10.1109/TKDE.2012.50}
}

@article{new_metrics2016cost,
  title={Cost-sensitive performance metric for comparing multiple ordinal classifiers},
  author={George, Nysia I and Lu, Tzu-Pin and Chang, Ching-Wei},
  journal={Art. Int.  Res.},
  volume={5},
  number={1},
  pages={135--143},
  year={2016}
  }

@article{new_metrics2019learning,
  title={Learning misclassification costs for imbalanced classification on gene expression data},
  author={Lu, Huijuan and Xu, Yige and Ye, Minchao and others},
  journal={BMC bioinformatics},
  volume={20},
  pages={1--10},
  year={2019},
  publisher={Springer},
  doi={10.1186/s12859-019-3255-x}
}

@inproceedings{new_metrics2019metric,
  title={Metric learning from imbalanced data},
  author={Gautheron, L{\'e}o and Habrard, Amaury and Morvant, Emilie and others},
  booktitle={2019 IEEE 31st Int. Conf. on Tools with Art. Int.  (ICTAI)},
  pages={923--930},
  year={2019},
  organization={IEEE}
}

@article{new_metrics2020imbalance,
  title={Imbalance accuracy metric for model selection in multi-class imbalance classification problems},
  author={Mortaz, Ebrahim},
  journal={Know.-Based Syst. },
  volume={210},
  pages={106490},
  year={2020},
  publisher={Elsevier},
  doi={10.1016/j.knosys.2020.106490}
}

@inproceedings{new_metrics2020top,
  title={Top-Rank-Focused Adaptive Vote Collection for the Evaluation of Domain-Specific Semantic Models},
  author={Lombardo, Pierangelo and Boiardi, Alessio and Colombo, Luca and others},
  booktitle={Proc. of the 2020 Conf. on Empirical Methods in Natural Language Proc.  (EMNLP)},
  pages={3081--3093},
  year={2020},
  doi={10.18653/v1/2020.emnlp-main.249}
}

@article{new_metrics2022extending,
  title={Extending {F1} metric, probabilistic approach},
  author={Sitarz, Mikolaj},
  journal={CoRR abs/2210.11997},
  year={2022},
  doi={10.48550/arXiv.2210.11997}  
}

@inproceedings{new_metrics2024f_1,
  title={Is {F1} Score Suboptimal for Cybersecurity Models? Introducing {Cscore}, a Cost-Aware Alternative for Model Assessment},
  author={Marwah, Manish and Narayanan, Asad and Jou, Stephen and others},
  booktitle={Conf. on Applied Mach. Learn. for Inf. Security},
  volume={3920},
  pages={190-209},
   year={2024},
   url={https://ceur-ws.org/Vol-3920/paper11.pdf}
}

@article{new_metrics2025standardization,
  title={Standardization of Weighted Ranking Correlation Coefficients},
  author={Lombardo, Pierangelo},
  journal={CoRR abs/2504.08428},
  year={2025},
  doi={10.48550/arXiv.2504.08428}
}

@inproceedings{cost_sensitive1998cost,
  title={Cost-sensitive learning with neural networks.},
  author={Kukar, Matjaz and Kononenko, Igor},
  booktitle={Proc. of the 13th European Conf. on Art. Int. (ECAI 98)},
  pages={445--449},
  year={1998}
}

@inproceedings{cost_sensitive1999metacost,
  title={{MetaCost}: A general method for making classifiers cost-sensitive},
  author={Domingos, Pedro},
  booktitle={Proc. of the 5th ACM SIGKDD Int. Conf. on Know. Disc. and Data Min.},
  pages={155--164},
  year={1999},
  url={https://dl.acm.org/doi/10.1145/312129.312220}
}

@article{cost_sensitive2000comparison,
  title={A comparison of selected artificial neural networks that help auditors evaluate client financial viability},
  author={Etheridge, Harlan L and Sriram, Ram S and Hsu, HY Kathy},
  journal={Decision Sciences},
  volume={31},
  number={2},
  pages={531--550},
  year={2000},
  publisher={Wiley Online Library},
  doi={10.1111/j.1540-5915.2000.tb01633.x}
}

@inproceedings{cost_sensitive2001foundations,
  title={The foundations of cost-sensitive learning},
  author={Elkan, Charles},
  booktitle={Int. joint Conf. on Art. Int.},
  volume={17},
  number={1},
  pages={973--978},
  year={2001}
}

@inproceedings{cost_sensitive2006influence,
  title={The influence of class imbalance on cost-sensitive learning: An empirical study},
  author={Liu, Xu-Ying and Zhou, Zhi-Hua},
  booktitle={6ht Int. Conf. on Data Min. (ICDM'06)},
  pages={970--974},
  year={2006},
  organization={IEEE},
  doi={10.1109/ICDM.2006.158}
}

@article{cost_sensitive2006test,
  title={Test strategies for cost-sensitive decision trees},
  author={Ling, Charles X and Sheng, Victor S and Yang, Qiang},
  journal={IEEE Trans. on Know. and Data Eng.},
  volume={18},
  number={8},
  pages={1055--1067},
  year={2006},
  publisher={IEEE},
  doi={10.1109/TKDE.2006.131}
}

@inproceedings{cost_sensitive2008cost,
  title={Cost-sensitive multi-class classification from probability estimates},
  author={O'Brien, Deirdre B and Gupta, Maya R and Gray, Robert M},
  booktitle={Proc. of the 25th Int. Conf. on Mach. Learn. (ICML)},
  pages={712--719},
  year={2008},
  doi={10.1145/1390156.139024}
}

@article{cost_sensitive2014imbalanced,
  title={Imbalanced dataset classification and solutions: a review},
  author={Ramyachitra, D and Manikandan, Parasuraman},
  journal={Int. Jour. of Comp. and Business Res. (IJCBR)},
  volume={5},
  number={4},
  pages={1--29},
  year={2014},
  issn={2229-6166}
}

@article{cost_sensitive2019example,
  title={Example-dependent cost-sensitive adaptive boosting},
  author={Zelenkov, Yuri},
  journal={Exp. Syst. Appl.},
  volume={135},
  pages={71--82},
  year={2019},
  doi = {10.1016/j.eswa.2019.06.009},
  publisher={Elsevier}
}

@article{cost_sensitive2021cost,
  title={A cost-sensitive deep learning-based approach for network traffic classification},
  author={Telikani, Akbar and Gandomi, Amir H and Choo, Kim-Kwang Raymond and others},
  journal={IEEE Trans. on Network and Service Manag.},
  volume={19},
  number={1},
  pages={661--670},
  year={2022},
  doi={10.1109/TNSM.2021.3112283}
}

@article{cost_sensitive2021multi,
  title={Multi-class misclassification cost matrix for credit ratings in peer-to-peer lending},
  author={Wang, Haomin and Kou, Gang and Peng, Yi},
  journal={Jour. of the Operational Res. Society},
  volume={72},
  number={4},
  pages={923--934},
  year={2021},
  publisher={Taylor \& Francis},
  doi={10.1080/01605682.2019.1705193}
}

@article{cost_sensitive2022cost,
  title={Cost-sensitive learning based on performance metric for imbalanced data},
  author={Aurelio, Yuri Sousa and de Almeida, Gustavo Matheus and de Castro, Cristiano Leite and others},
  journal={Neural Proc.  Lett.},
  volume={54},
  number={4},
  pages={3097--3114},
  year={2022},
  publisher={Springer},
  doi={10.1007/s11063-022-10756-2}
}

@inproceedings{diabetes2007using,
  title={Using Asymmetric Classification Cost Matrices in Predicting Diabetes},
  author={Ghosh, Bishwadip and Hasley, Joseph},
  booktitle={ICDSS 2007 Proceedings},
  year={2007}
}

@inproceedings{diabetes2020prediction,
  title={Prediction of diabetes using cost sensitive learning and oversampling techniques on Bangladeshi and Indian female patients},
  author={Pranto, Badiuzzaman and Mehnaz, Sk Maliha and Momen, Sifat and others},
  booktitle={2020 5th Int. Conf. on Inf. Tech. Res. (ICITR)},
  pages={1--6},
  year={2020},
  organization={IEEE},
  doi={10.1109/ICITR51448.2020.9310892}
}

@article{cost_sensitive1977zetatm,
title = {{ZETATM} analysis {A} new model to identify bankruptcy risk of corporations},
journal = {Jour. of Banking \& Finance},
author = {Edward I. Altman and Robert G. Haldeman and P. Narayanan},
volume = {1},
number = {1},
pages = {29-54},
year = {1977},
issn = {0378-4266},
doi = {https://doi.org/10.1016/0378-4266(77)90017-6}
}

@article{cost_matrix2012classification,
  title={Classification cost: An empirical comparison among traditional classifier, Cost-Sensitive Classifier, and {MetaCost}},
  author={Kim, Jungeun and Choi, Keunho and Kim, Gunwoo and others},
  journal={Exp. Syst.  with Appl.},
  volume={39},
  number={4},
  pages={4013--4019},
  year={2012},
  publisher={Elsevier},
  doi={doi.org/10.1016/j.eswa.2011.09.071}
}

@article{spearman1961proof,
  title={The proof and measurement of association between two things.},
  author={Spearman, Charles},
  journal={The American Jour. of Psychology},
  year= {1904},
  volume={15}, 
  number={1},
  pages={72-101},
  publisher={AUniversity of Illinois Press}
}

@inproceedings{credit2014example,
  title={Example-dependent cost-sensitive logistic regression for credit scoring},
  author={Bahnsen, Alejandro Correa and Aouada, Djamia and Ottersten, Bj{\"o}rn},
  booktitle={2014 13th International conference on machine learning and applications},
  pages={263--269},
  year={2014},
  organization={IEEE}
}

\end{document}